\documentclass[10pt,twocolumn,letterpaper]{article}

\usepackage{cvpr}              
\usepackage{multirow}

\usepackage[dvipsnames]{xcolor}

\definecolor{cvprblue}{rgb}{0.21,0.49,0.74}
\usepackage[pagebackref,breaklinks,colorlinks,citecolor=cvprblue]{hyperref}

\def\paperID{11539} 
\def\confName{CVPR}
\def\confYear{2024}
 
\title{PatchCraft: Exploring Texture Patch for Efficient AI-generated Image Detection }

\author{
Nan Zhong \ \ \ 
Yiran Xu \ \ \ 
Sheng Li \ \ \
Zhenxing Qian \ \ \ 
Xinpeng Zhang \ \ \ 
 \\[2mm]
{Fudan University}
\\[2mm]
{\{nzhong20, yrxu23, lisheng, zxqian, zhangxinpeng\}@fudan.edu.cn}
}

\begin{document}


\twocolumn[{%
\maketitle
\begin{center}
  \centering
  \includegraphics[width=7in,clip]{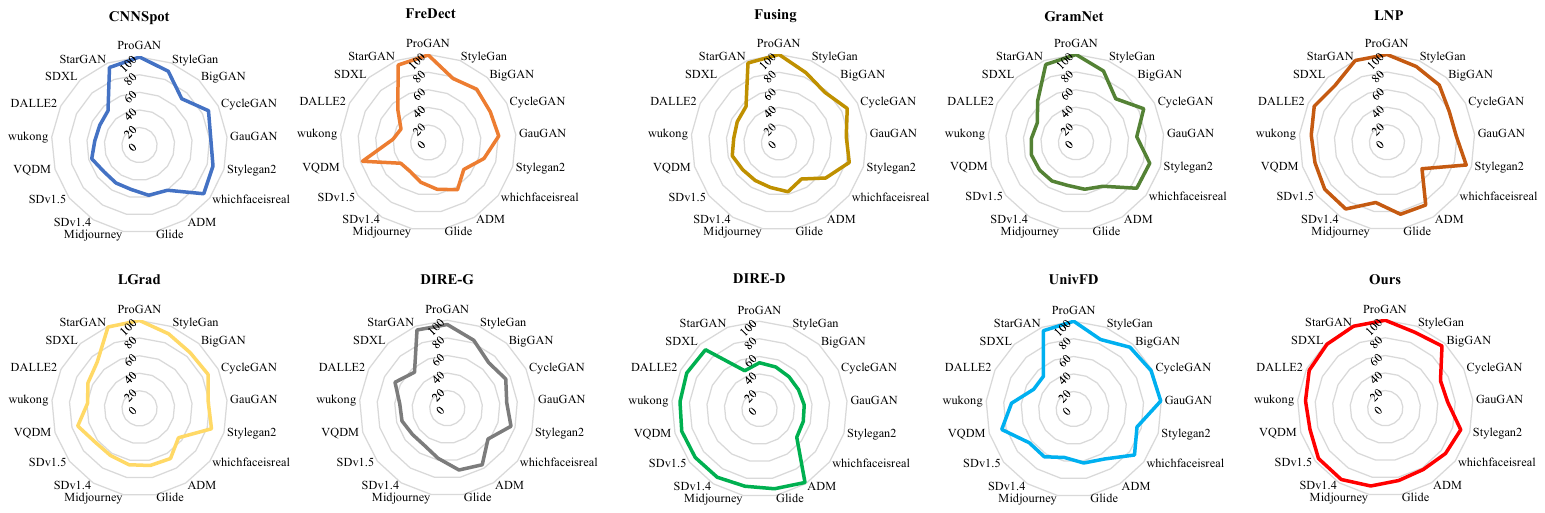}
  \captionof{figure}{We conduct a comprehensive AI-generated image detection benchmark, including 16 kinds of prevalent generative models \cite{karras2017progressive,karras2019style,brock2018large,zhu2017unpaired,choi2018stargan,park2019semantic,karras2020analyzing,dhariwal2021diffusion,nichol2021glide,rombach2022high,gu2022vector} and commercial APIs \cite{wukong, dall-e-2} like Midjourney \cite{midjourney}. Multiple cutting-edge detectors \cite{wang2020cnn,frank2020leveraging,ju2022fusing,liu2020global,liu2022detecting,tan2023learning,wang2023dire,ojha2023towards} are presented in the benchmark. We visualize the results with a radar chart. The concentric circles denote the detection accuracy. Our approach outperforms the state-of-the-art detector about 4\% over average detection accuracy.}
\end{center}%
}]

\begin{abstract}
Recent generative models show impressive performance in generating photographic images. Humans can hardly distinguish such incredibly realistic-looking AI-generated images from real ones. AI-generated images may lead to ubiquitous disinformation dissemination. Therefore, it is of utmost urgency to develop a detector to identify AI-generated images. Most existing detectors suffer from sharp performance drops over unseen generative models. In this paper, we propose a novel AI-generated image detector capable of identifying fake images created by a wide range of generative models. We observe that the texture patches of images tend to reveal more traces left by generative models compared to the global semantic information of the images. A novel {\it Smash$\&$Reconstruction} preprocessing is proposed to erase the global semantic information and enhance texture patches. Furthermore, pixels in rich texture regions exhibit more significant fluctuations than those in poor texture regions. Synthesizing realistic rich texture regions proves to be more challenging for existing generative models. Based on this principle, we leverage the inter-pixel correlation contrast between rich and poor texture regions within an image to further boost the detection performance. 
In addition, we build a comprehensive AI-generated image detection benchmark, which includes 17 kinds of prevalent generative models, to evaluate the effectiveness of existing baselines and our approach. Our benchmark provides a leaderboard for follow-up studies. Extensive experimental results show that our approach outperforms state-of-the-art baselines by a significant margin. Our project: https://fdmas.github.io/AIGCDetect/

\end{abstract}

\section{Introduction}

Over the past years, generative models have achieved remarkable progress. Various models, including VAE \cite{kingma2013auto}, GAN \cite{goodfellow2020generative}, GLOW \cite{kingma2018glow}, Diffusion \cite{sohl2015deep} and their variations \cite{brock2018large,karras2017progressive,karras2019style,ho2020denoising,song2019generative,rombach2022high}, are flourishing in the image composition domain. At the end of 2017, DeepFake videos, including face forgery generated by GANs, attracted great attention in academic and societal circles \cite{suwajanakorn2017synthesizing,arik2018neural,kim2018deep}. From 2021, diffusion-based models \cite{dhariwal2021diffusion,nichol2021glide,rombach2022high,gu2022vector} have become a new paradigm in the image composition domain, generating high-quality facial images and arbitrary semantic images. Especially, recent text-to-image models, such as Stable Diffusion \cite{rombach2022high}, generate arbitrary fake images by text descriptions. Commercial APIs like Midjourney \cite{midjourney}, DALL-E 2 \cite{dall-e-2}, \etc., allow users to obtain fake photographic images without requisite expertise. With the rapid advancement of generative models, fake images become more and more realistic-looking and visually comprehensible. The easy accessibility of fake images may intensify concerns regarding the ubiquitous dissemination of disinformation. For example, the nefarious usage of AI-generated images has been confirmed, with a fake image depicting an explosion at the Pentagon being widely circulated on Twitter \cite{news}. Therefore, it is of utmost urgency to develop an effective detector capable of identifying fake images generated by modern generative models. Note that fake images can be roughly grouped into two categories, \ie, AI-generated images and manipulated-based images. Specifically, most AI-generated fake images are created by unconditional generative models (mapping a random noise to a real image) or conditional generative models like text2image models \cite{rombach2022high}. There are no corresponding exact real images for AI-generated images. Manipulated-based fake images like the famous Deepfake \cite{rossler2019faceforensics++} are created by tampering real images. In this paper, we focus on the AI-generated fake image detection. 

Numerous detectors have been proposed to identify AI-generated images  \cite{gragnaniello2021gan,barni2020cnn,frank2020leveraging,wang2020cnn,ju2022fusing,liu2022detecting,tan2023learning,liu2020global,wang2023dire}. Before the prevalence of diffusion-based models, most researchers focus on detecting GAN-based fake images \cite{liu2020global,frank2020leveraging,wang2020cnn,ju2022fusing,liu2022detecting,barni2020cnn}. Some detectors are dedicated to identifying fake facial images based on specific facial features \cite{liu2020global,cao2022end,hulzebosch2020detecting,pu2022learning,rossler2019faceforensics++,liu2021spatial}, and others handle fake images with arbitrary categories \cite{gragnaniello2021gan,frank2020leveraging,wang2020cnn,ju2022fusing,liu2022detecting,tan2023learning}. It is challenging for a classifier trained over a certain type of generator (like ProGAN \cite{karras2017progressive}) to work effectively over fake images from unseen sources (like BigGAN \cite{brock2018large}). In open-world applications, fake images always come from various sources, such as being generated by unknown approaches. Therefore, the fake image classifier requires generalization across various generative models. In previous studies, researchers bolster the generalization of detectors based on different views like global textures analysis \cite{liu2020global}, frequency-level artifacts \cite{frank2020leveraging}, data augmentation \cite{wang2020cnn}, \etc. Nowadays, diffusion-based generators, which synthesize more high-quality images than GAN-based images, unleash a new wave of photographic image creation. Some detectors \cite{wang2023dire,sha2022fake} are dedicated to identify diffusion-generated images based on the artifacts inevitably left by diffusion models.

In this paper, we design an AI-generated fake image detector capable of identifying various unseen source data, including GAN-based \cite{karras2017progressive,karras2019style,brock2018large,zhu2017unpaired,zhu2017unpaired,choi2018stargan,park2019semantic} and diffusion-based generative models \cite{dhariwal2021diffusion,nichol2021glide,rombach2022high,gu2022vector}. Specifically, we only train a classifier based on a training set generated by ProGAN \cite{karras2017progressive}, and it works effectively over various models like Stable Diffusion \cite{rombach2022high}, BigGAN \cite{brock2018large}, Midjourney \cite{midjourney}, \etc. To accomplish this goal, we need to extract a universal fingerprint across various generative models rather than directly training a binary classifier from the spatial domain. Specifically, we identify AI-generated images based on the inter-pixel correlation. The inter-pixel correlation is closely related to the high-frequency component of an image. Previous studies \cite{frank2020leveraging,chandrasegaran2021closer,durall2020watch} demonstrate that various generative models leave many artifacts in high-frequency components of synthetic images rather than in low-frequency components related to the global semantic information of images. Therefore, we propose {\it Smash$\&$Reconstruction} to break the global semantic information (suppress low-frequency components) and magnify artifacts left by generative models. It is difficult to identify synthetic images from their semantic information for the state-of-the-art generative models. As a result, the performance of detectors based on semantic information sharply degrades for realistic-looking fake images. {\it Smash$\&$Reconstruction} makes the subsequent classifier distinguish real or synthetic images by the inter-pixel correlation extracted from texture patches instead of the global semantic information.

Next, We recall the evolution of fake image composition. In the early stage, generative models can create high-quality facial images like CelebA set \cite{liu2018large}, but they falter in creating multiple categories images like ImageNet set \cite{deng2009imagenet}. The reason is that the distribution approximation based on random noise for facial images is much easier than that of multiple categories images. The entropy of the facial image distribution is much smaller than that of various categories images. It is difficult to approximate a distribution with high entropy from random noise. Modern state-of-the-art generative models can create realistic-looking fake images with arbitrary content. However, the poor texture regions of synthetic images still behave differently from rich texture regions. Generative models leave different artifacts between the poor and rich texture regions within synthetic images stemming from the entropy discrepancy (distribution approximation difficulty) of the two regions. We find that this characteristic of synthetic images widely exists across various cutting-edge generative models. Accordingly, we design an AI-generated fake image detector which works effectively across various generative models.

The main contributions can be summarized as follows.

\begin{itemize}
    \item We propose {\it Smash$\&$Reconstruction} to break the semantic information of images and enhance the texture regions of images. {\it Smash$\&$Reconstruction} makes the detector focus on the inter-pixel correlation of images and significantly boosts the generalization of the detector.
	\item We propose a universal fingerprint that widely exists in various AI-generated images, including GAN-based or diffusion-based generative models. The universal fingerprint of synthetic images is based on the inter-pixel correlation contrast between rich and poor texture regions within an image.
    \item We conduct a comprehensive AI-generated image detection benchmark including various remarkable generative models. It is convenient for follow-up studies to compare their performance with the same condition. Extensive experimental results show that our approach outperforms the state-of-the-art detector with a clear margin. 
\end{itemize}

\section{Related Work}

\noindent\textbf{Image generation} has achieved a series of remarkable progress in recent years, which aims to generate photographic images from random noise  \cite{karras2017progressive,karras2019style,brock2018large,zhu2017unpaired,choi2018stargan,park2019semantic,dhariwal2021diffusion} or text description  \cite{nichol2021glide,gu2022vector,reed2016generative,qiao2019mirrorgan}. We denote the distribution of real images as $p(x)$. It is difficult to give a concrete formulaic expression of $p(x)$. In the image generation field, researchers aim to adopt $g_{\theta}(z)$ to approximate $p(x)$ as closely as possible, where $g_{\theta}(\cdot)$ and $z$ denote a learnable function (deep model) and random noise, respectively. For instance, famous GANs  \cite{karras2017progressive,karras2019style,brock2018large,zhu2017unpaired,choi2018stargan,park2019semantic} and their variants employ a generator to transfer random noise $z$ (sampled from an isotropic Gaussian distribution) to fake images, that is, mapping a Gaussian distribution into an image distribution. Then, they employ a discriminator, which aims to identify whether an image is real or fake, to measure the distance between the synthetic image distribution $g_{\theta}(z)$ and real image $p(x)$. The entire process of GANs consists of two steps, \ie,, 1) measuring the distance between synthetic image distribution $g_{\theta}(z)$ and real image $p(x)$; 2) minimizing their distance by updating the generator $g_{\theta}(\cdot)$. In terms of prevalent diffusion models \cite{rombach2022high,nichol2021glide,dhariwal2021diffusion}, they gradually introduce Gaussian noise to a real image, ultimately transforming it into an isotropic Gaussian noise. Subsequently, they learn to reverse the diffusion process in order to generate an image from a random noise sampled from an isotropic Gaussian distribution. In a nutshell, various image generation algorithms aim to approximate real image distribution $p(x)$ from an isotropic Gaussian distribution.

\noindent\textbf{AI-generated fake image detection} aims to distinguish whether an image is real (generated by cameras or phones) or fake (generated by generative models). Prior studies \cite{wang2020cnn,frank2020leveraging,ju2022fusing,liu2020global,tan2023learning,wang2023dire} treat fake image detection as a binary classification task. A deep model-based classifier (like ResNet-50) always performs well when the fake images of the test and training set are generated by the same generative model. However, its performance significantly degrades for unseen generative models. To tackle the generalization of the detector, many researchers have designed various approaches to explore a universal fingerprint across different generative models. Frank \etal. \cite{frank2020leveraging} observe consistently anomalous behavior of fake images in the frequency domain and train the detector in the frequency domain. Durall \etal. \cite{durall2020watch} further explain that the upsampling operation results in the frequency abnormality of fake images. Liu \etal. \cite{liu2022detecting} find that the noise pattern of an image in the frequency domain can be used as a fingerprint to improve the generalization of the detector. Liu \etal. \cite{liu2020global} focus on the texture of images and incorporate the Gram Matrix extraction into the typical ResNet structure to conduct fake image detection. Wang \etal. \cite{wang2020cnn} propose that easy data augmentation can significantly boost the generalization for the detector. More recently, Wang \etal. \cite{wang2023dire} find that fake images generated by a diffusion model are more likely to be accurately reconstructed by a pre-trained diffusion model. Therefore, they adopt the reconstruction error of an image as the fingerprint to identify diffusion-generated fake images. 

\begin{figure}[]
   \centering
   \includegraphics[width=3.2in,clip,trim= 230 190 330 0]{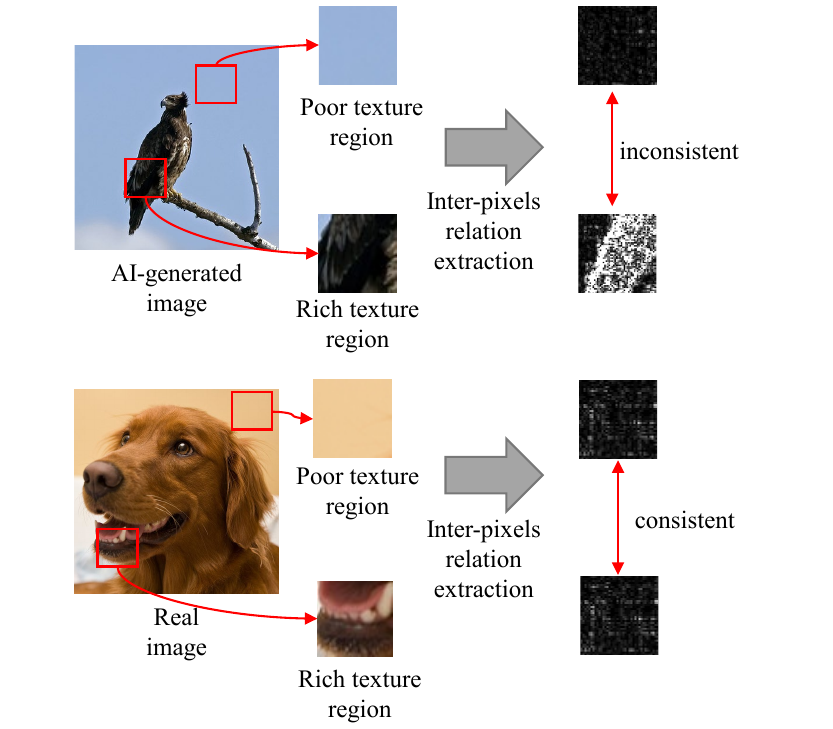}
   \caption{The illustration of our motivation.}
   \label{fig_motivation}
\end{figure}

\begin{figure*}[]
   \centering
   \includegraphics[width=6.5in,clip,trim= 0 250 0 0]{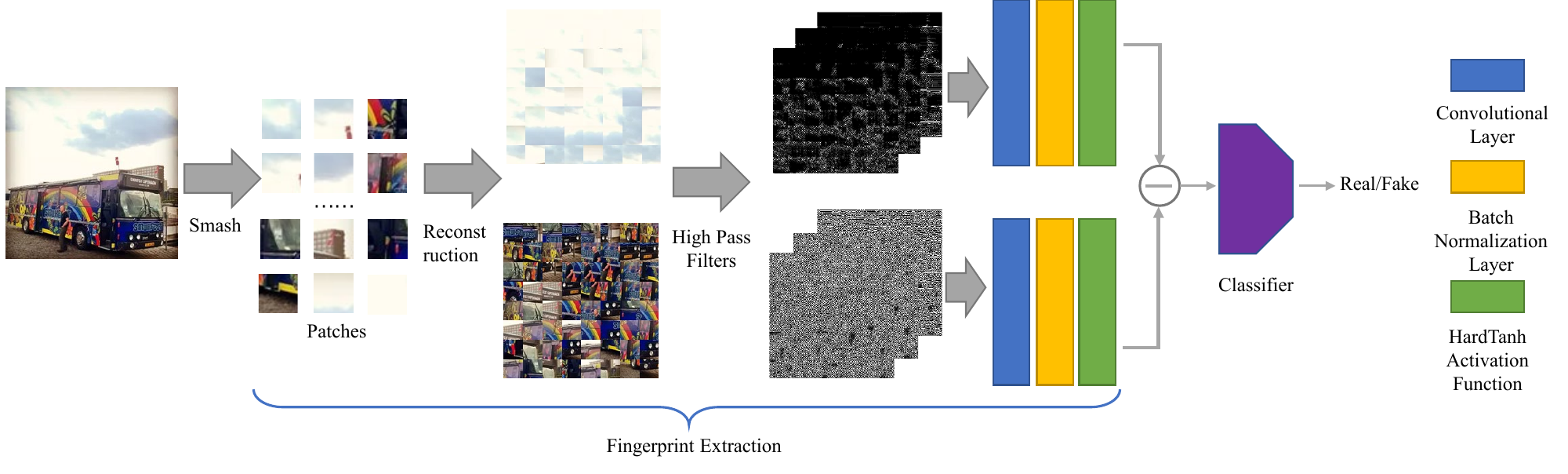}
   \caption{The framework of our approach.}
   \label{fig_framework}
\end{figure*}

\section{Proposed Method}

\subsection{Motivation}
In this section, we aim to answer "How to identify AI-generated images?". A naive detection approach is training a common classifier like ResNet-50 to distinguish real and synthetic images directly. However, the performance of naive classifier always significantly degrades for unseen generative models. The reason is that the classifier largely relies on the global semantic information of images to make decisions, that is, the naive classifier can not catch the universal artifacts features of various generative models. Previous studies show that generative models leave artifacts in the high-frequency components due to the upsampling operation \cite{durall2020watch}. The abnormality in high-frequency components can result in the inter-pixel correlation discrepancy between synthetic and real images. Based on the image device/tracking forensics  \cite{lukas2006digital}, inter-pixel correlation (also is similar with PRNU noise pattern  \cite{chierchia2014bayesian,scherhag2019detection}) is determined by its camera device (CMOS) and ISP (Image Signal Processing). Although some cutting-edge generative models create impressive images from the semantic view, they still struggle to simulate inter-pixel correlation of real images. Therefore, we propose {\it Smash$\&$Reconstruction} to break the global semantic information and make the classifier focus on the  inter-pixel correlations of images.

The goal of the image generation is to use an isotropic Gaussian distribution to approximate the real image distribution $p(x)$. The approximation difficulty depends on the entropy of the $p(x)$, which is related to the diversity of the real image $p(x)$. It is more difficult to approximate $p(x)$ with high diversity. This proposition is also corroborated by the development of the image generation field. In the early stage, generative models can create high-quality facial image, which is hardly distinguished from real facial images by humans, while their performance significantly drops when the training set is changed to a diversity set like ImageNet \cite{deng2009imagenet}. The real image distribution of ImageNet \cite{deng2009imagenet} (large entropy) is much more diverse than that of a facial set like CelebA \cite{liu2018large}. 

Although existing state-of-the-art generative models can create realistic fake images with various content, they leave different artifacts in the rich or poor texture regions. Rich texture regions of an image behave more diversely than poor texture regions. As a result, it is more difficult for generative models to approximate rich texture regions of real images. For a real image, the inter-pixel correlations, which are determined by its camera device (CMOS) and ISP (Image Signal Processing), between rich and poor texture regions are very close. Therefore, we leverage the inter-pixel correlation contrast between rich and poor texture regions of an image as a fingerprint to identify AI-generated fake images. Fig. \ref{fig_motivation} illustrates our motivation of the inter-pixel correlation contrast.

The cornerstone of the generalization of the detector relies on the fingerprint feature extraction. In prior studies \cite{frank2020leveraging, ju2022fusing, liu2022detecting, tan2023learning, wang2023dire}, researchers designed various fingerprint features. For instance, Liu \etal. \cite{liu2022detecting} adopt the noise pattern extracted by a well-trained denoising network as a fingerprint feature for images. Intuitively, if the fingerprint feature exists across various generative models including GANs, diffusion and their variants, the detector gains great generalization. We adopt the inter-pixel correlation contrast between rich and poor texture regions as the fingerprint feature. Since our fingerprint feature is based on the inherent weakness of distribution approximation, the detector works effectively over fake images that have not been encountered during the training phase.

\subsection{Fingerprint Feature Extraction}

Fig. \ref{fig_framework} shows the framework of our approach, which consists of two steps, \ie,, fingerprint feature extraction and detector training. Since our fingerprint feature is the inter-pixel correlation contrast between rich and poor texture regions, we aim to suppress the semantic information of images and extract inter-pixel correlation. Each input image is transferred into two parts, \ie., rich texture regions and poor texture regions. We first randomly crop the input image into multiple patches and sort patches based on their texture diversity. Then, we adopt these patches to reconstruct two images consisting of rich or poor texture patches, respectively. We name this process as {\it Smash$\&$Reconstruction}, which is used to suppress the semantic information of the images. {\it Smash$\&$Reconstruction} makes the detector distinguish real or fake images regardless of the semantic information of images. If the detector distinguishes real or fake images based on semantic information, its performance significantly drops for fake images whose semantic information is almost impeccable. The texture diversity can be measured by pixel fluctuation degree. Compared with rich texture regions, pixels located in the poor texture regions are prone to the same value. For a $M\times M$ size patch, we measure its texture diversity by summing the residual of four directions including diagonal, counter-diagonal, horizontal direction and vertical direction. Equation \ref{e1} gives the concrete expression,
\begin{equation}\label{e1}
\small
\begin{aligned}
l_{ div } = \sum_{i = 1}^{M} \sum_{j = 1}^{M-1}(\left | x_{i,j} - x_{i,j+1} \right | )  + \sum_{i = 1}^{M-1} \sum_{j = 1}^{M}(\left | x_{i,j} - x_{i+1,j} \right | 
    \\ + \sum_{i = 1}^{M-1} \sum_{j = 1}^{M-1}(\left | x_{i,j} - x_{i+1,j+1} \right |  + \sum_{i = 1}^{M-1} \sum_{j = 1}^{M-1}(\left | x_{i+1,j} - x_{i,j+1} \right |
). 
\end{aligned}
\end{equation}
Patch images with rich texture are assigned to a large $l_{div}$. As Fig. \ref{fig_framework} shows, the rich texture part consists of patches with high $l_{div}$, while the poor texture part consists of patches with low $l_{div}$. The semantic information of both two parts is broken due to {\it Smash$\&$Reconstruction}.

Furthermore, we adopt a set of high-pass filters proposed in SRM \cite{fridrich2012rich} (Spatial Rich Model) to extract the noise pattern of two parts. High-pass filters are conducive to suppressing semantic information and magnifying the inter-pixel correlation, which can be used as a fingerprint across various fake images. Previous studies also demonstrate that fake images behave abnormally in high frequency \cite{frank2020leveraging, durall2020watch}. We adopt 30 different high-pass filters, and their concrete parameters can be found in the supplement. Subsequently, we add a learnable convolution block consisting of one convolution layer, one batch normalization layer, and one Hardtanh activation function after the outputs of high-pass filters. Our fingerprint of an image, which denotes the inter-pixel correlation contrast between rich and poor texture regions, is measured by the residual of the outputs of the learnable convolution block. 

An alternative advantage of our approach is that the fingerprint feature extraction is robust to the size of the input image. During the common binary classifier training process, if the size of the input image is too large, users have to reduce the size of the images by downsampling. However, since the artifacts left by generative models are very weak, the downsampling operation inevitably further erases the artifacts, leading to the performance degradation of the detector. In our scheme, the size of the input image is irrelevant to the dimension of the fingerprint, which only depends on the patch size and the number of the patch.  

As Fig. \ref{fig_framework} shows, some boundaries, also named blocking artifacts, exist across various patches. We adopt high-pass filters to process the whole rich/poor texture reconstructed images, that is, high-pass filters also sweep boundaries between patches. In the rich/poor texture reconstructed images, each patch is sorted from top left to bottom right based on their diversity. The patch located in the top-left corner contains the poorest/richest texture. We retain the boundaries of patches in our approach. Then, we leverage high-pass filters across boundaries to extract variation trends of patch diversity, which is conducive to improving the performance of our detector.

\subsection{AI-generated Fake Image Detection}

We extract the fingerprint of each input image and feed it into a binary classifier to determine whether the input image is real or fake. In light of the large gap in the fingerprint features between real and fake images, even a simple structure classifier consisting of cascaded convolution blocks can effectively distinguish real or fake images. Therefore, our classifier is made up of some common cascaded convolution blocks. The details of the classifier can be found in the supplement. Artifacts left by generative models are very weak, especially in some impressive generative models. Pooling layers reduce the size of feature maps, resulting in artifacts destruction. However, pooling layers are inevitable in common classifiers to mitigate computation. Therefore, we require sufficient convolution blocks before the first pooling layers. We adopt four convolution blocks consisting of one convolution layer, one batch normalization layer and one Relu activation function before the first average pooling layer.  

We denote the fingerprint feature extraction process as $T(\cdot)$, and we aim to minimize the cross-entropy loss $l_{cle}$ by updating the parameters of the classifier $f_{\theta}(\cdot)$ and the learnable convolution block adopted in fingerprint feature module. We give the formulaic expression as follows,
\begin{equation}
\small
\begin{aligned}
l_{cle} = - \sum_{i=1}^{N}(y_ilog(f_{\theta}(T(x_i)))  +  (1-y_i)log(1-f_{\theta}(T(x_i)) ,
\end{aligned}
\end{equation}
where $x_i$ and $y_i$ denote the input image and its corresponding label. The size of the patch and the number of patches used to construct fingerprint features are fixed in advance. In the inference/test phase, we first extract the fingerprint feature from a suspicious image with an arbitrary size, and the classifier returns the result for the fingerprint feature.

\section{Experimental Results}

In this section, we give a comprehensive evaluation of our approach with various state-of-the-art generative models. We also employ multiple baselines to demonstrate the superiority of the proposed approach. 

\subsection{Experimental Setup}

\begin{table}[htbp]
  \centering
  
        \begin{tabular}{cccc}
        \toprule
    Generative model & Size  & Number & Source \\
    \hline
    ProGAN \cite{karras2017progressive} & 256×256 & 8.0k  & LSUN \\
    StyleGAN \cite{karras2019style} & 256×256 & 12.0k  & LSUN \\
    BigGAN \cite{brock2018large} & 256×256 & 4.0k  & ImageNet \\
    CycleGAN \cite{zhu2017unpaired} & 256×256 & 2.6k  & ImageNet \\
    StarGAN \cite{choi2018stargan} & 256×256 & 4.0k  & CelebA \\
    GauGAN \cite{park2019semantic} & 256×256 & 10.0k & COCO \\
    Stylegan2 \cite{karras2020analyzing} & 256×256 & 15.9k & LSUN \\
    WFIR \cite{whichfaceisreal} & 1024×1024 & 2.0k  & FFHQ \\
    ADM \cite{dhariwal2021diffusion}  & 256×256 & 12.0k & ImageNet \\
    Glide \cite{nichol2021glide} & 256×256 & 12.0k & ImageNet \\
    Midjourney \cite{midjourney} & 1024×1024 & 12.0k & ImageNet \\
    SDv1.4 \cite{rombach2022high} & 512×512 & 12.0k & ImageNet \\
    SDv1.5 \cite{rombach2022high} & 512×512 & 16.0k & ImageNet \\
    VQDM \cite{gu2022vector} & 256×256 & 12.0k & ImageNet \\
    Wukong \cite{wukong} & 512×512 & 12.0k & ImageNet \\
    DALL-E 2 \cite{dall-e-2} & 256×256 & 2.0k    & ImageNet \\
    SDXL \cite{podell2023sdxl}  & 1024×1024 & 4.0k    & COCO \\
    \bottomrule
    \end{tabular}%
    \caption{The description of generative models. SD and WFIR denote Stable Diffusion and whichfaceisreal, respectively.}
  \label{tab_dataset}%
\end{table}%

\begin{table*}[]
  \centering
  
    \begin{tabular}{ccccccccccc}
    \toprule
    Generator & CNNSpot & FreDect & Fusing & GramNet & LNP   & LGrad & DIRE-G & DIRE-D & UnivFD & Ours \\
    \hline
    ProGAN & \textbf{100.00} & 99.36 & \textbf{100.00} & 99.99 & 99.95 & 99.83 & 95.19 & 52.75 & 99.81 & \textbf{100.00} \\
    StyleGan & 90.17 & 78.02 & 85.20 & 87.05 & \underline{ 92.64} & 91.08 & 83.03 & 51.31 & 84.93 & \textbf{92.77} \\
    BigGAN & 71.17 & 81.97 & 77.40 & 67.33 & 88.43 & 85.62 & 70.12 & 49.70 & \underline{ 95.08} & \textbf{95.80} \\
    CycleGAN & \underline{ 87.62} & 78.77 & 87.00 & 86.07 & 79.07 & 86.94 & 74.19 & 49.58 & \textbf{98.33} & 70.17 \\
    StarGAN & 94.60 & 94.62 & 97.00 & 95.05 & \textbf{100.00} & 99.27 & 95.47 & 46.72 & 95.75 & \underline{99.97} \\
    GauGAN & \underline{ 81.42} & 80.57 & 77.00 & 69.35 & 79.17 & 78.46 & 67.79 & 51.23 & \textbf{99.47} & 71.58 \\
    Stylegan2 & 86.91 & 66.19 & 83.30 & 87.28 & \textbf{93.82} & 85.32 & 75.31 & 51.72 & 74.96 & \underline{ 89.55} \\
    whichfaceisreal & \textbf{91.65} & 50.75 & 66.80 & 86.80 & 50.00 & 55.70 & 58.05 & 53.30 & \underline{ 86.90} & 85.80 \\
    ADM & 60.39 & 63.42 & 49.00 & 58.61 & \underline{ 83.91} & 67.15 & 75.78 & \textbf{98.25} & 66.87 & 82.17 \\
    Glide & 58.07 & 54.13 & 57.20 & 54.50 & 83.50 & 66.11 & 71.75 & \textbf{92.42} & 62.46 & \underline{ 83.79} \\
    Midjourney & 51.39 & 45.87 & 52.20 & 50.02 & 69.55 & 65.35 & 58.01 & \underline{ 89.45} & 56.13 & \textbf{90.12} \\
    SDv1.4 & 50.57 & 38.79 & 51.00 & 51.70 & 89.33 & 63.02 & 49.74 & \underline{ 91.24} & 63.66 & \textbf{95.38} \\
    SDv1.5 & 50.53 & 39.21 & 51.40 & 52.16 & 88.81 & 63.67 & 49.83 & \underline{ 91.63} & 63.49 & \textbf{95.30} \\
    VQDM & 56.46 & 77.80 & 55.10 & 52.86 & 85.03 & 72.99 & 53.68 & \textbf{91.90} & 85.31 & \underline{ 88.91} \\
    wukong & 51.03 & 40.30 & 51.70 & 50.76 & 86.39 & 59.55 & 54.46 & \underline{ 90.90} & 70.93 & \textbf{91.07} \\
    DALLE2 & 50.45 & 34.70 & 52.80 & 49.25 & 92.45 & 65.45 & 66.48 & \underline{ 92.45} & 50.75 & \textbf{96.60} \\
    SDXL & 53.03 & 51.23 & 55.60 & 64.53 & 87.75 & 71.30 & 55.35 & \underline{ 91.28} & 50.73 & \textbf{98.43} \\
    \hline
    Average & 69.73 & 63.28 & 67.63 & 68.43 & \underline{ 85.28} & 75.11 & 67.90 & 72.70 & 76.80 & \textbf{89.85}\\
    \bottomrule
    \end{tabular}%
    \caption{The detection accuracy comparison between our approach and baselines. DIRE-D denotes DIRE detector trained over fake images from ADM, and we adopt the checkpoint released by their official codes.  DIRE-G denotes DIRE detector trained over the same training set (ProGAN) as others. Among all detectors, the best result and the second-best result are denoted in boldface and underlined, respectively.}
  \label{tab_main}%
\end{table*}%


\noindent\textbf{Datasets.} In order to comprehensively evaluate our approach, we adopt 17 different generative models. The details of test datasets are shown in Table \ref{tab_dataset}. The fake images generated by GAN-based generative models and whichfaceisreal (WFIR) \cite{whichfaceisreal} are contributed by CNNSpot \cite{wang2020cnn}. Images of DALL-E 2 \cite{dall-e-2} and SDXL \cite{podell2023sdxl} are curated by ours. Other images are contributed by GenImage \cite{zhu2023genimage}. We use the label of the image as the prompt to generate fake images for some text2image generative models like DALL-E 2 and Midjourney. In terms of the training set, we adopt the training set proposed by CNNSpot \cite{wang2020cnn}, which is also widely used in previous studies \cite{ojha2023towards,tan2023learning}. The training set contains 720k images made up of 360k real images from LSUN \cite{yu2015lsun} and 360k fake images from ProGAN \cite{karras2017progressive}. Since the detector does NOT know the generative model of fake images in the practical application, the generalization, which is the performance of the detector for unseen source data during the training phase, is an important criterion for the detector evaluation. As Table \ref{tab_dataset} shows, we adopt various generative models, including some advanced commercial APIs like Midjourney, to demonstrate the generalization of the proposed approach. Compared with DeepFake detection, which is dedicated to identifying face manipulation, we aim to identify AI-generated images with various semantic content. In other words, we focus on AI-synthesized image detection rather than AI-manipulated image detection. In terms of AI-generated facial images, We also evaluate detectors over facial images like whichfaceisreal (WFIR) \cite{whichfaceisreal} and StarGAN \cite{choi2018stargan}. It is a challenging task that our training set does not contain facial images.

\noindent\textbf{Baselines} 1) CNNSpot (CVPR'2020) \cite{wang2020cnn} propose a simple yet effective fake image detector. They observe that data augmentation, including JPEG compression and Gaussian blur, can boost the generalization of the detector and adopt ResNet-50 as a classifier. 2) FreDect (ICML'2020) \cite{frank2020leveraging} proposes the frequency abnormality of fake images and conducts fake image detection from the frequency domain. 3) Fusing (ICIP'2022) \cite{ju2022fusing} combines the patch and global information of images to train the classifier. 4) GramNet (CVPR'2020) \cite{liu2020global} aims to improve the generalization of the detector by incorporating a global texture extraction into the common ResNet structure. 5) LNP (ECCV'2022) \cite{liu2022detecting} extracts the noise pattern of spatial images based on a well-trained denoising model. Then, it identifies fake images from the frequency domain of the noise pattern. 6) LGrad (CVPR'2023) \cite{tan2023learning} extracts gradient map, which is obtained by a well-trained image classifier, as the fingerprint of an image, and conducts a binary classification task based on gradient maps. 7) DIRE (ICCV'2023) \cite{wang2023dire} is dedicated to identifying fake images generated by diffusion-based images. It leverages the reconstruction error of the well-trained diffusion model as a fingerprint. 8) UnivFD (CVPR'2023) \cite{ojha2023towards} uses a feature space extracted by a large pre-trained vision-language model \cite{radford2021learning} to train the detector. The large pre-trained model leads to a smooth decision boundary, which improves the generalization of the detector.

For a fair comparison of the generalization, all baselines (except for DIRE-D) are trained over the same training set as our approach (360k real images from LSUN and 360k fake images generated by ProGAN). DIRE-D is a pre-trained detector trained over ADM dataset and its checkpoint is provided by their official codes.

\noindent\textbf{Details for Our approach} In our approach, we first smash spatial images into multiple patches to break the semantic information of input images. The number of patches and the size of each patch are set as 192 and $32\times32$, respectively. We sort patches based on their diversity. Subsequently, we divide these patches into two parts, i.e., rich texture patches with top 33\% diversity (64 patches) and poor texture patches with bottom 33\% diversity. The size of the reconstructed image made up of 64 patches is $256\times256$. We adopt Adam optimizer with 0.001 learning rate to update parameters. The batch size is 32. We adopt three data augmentations, including JPEG compression (QF$\sim$ Uniform[70, 100]), Gaussian blur $\sigma\sim$ Uniform[0, 1]), and Downsampling ($r\sim$ Uniform[0.25, 0.5]), to improve the robustness of our approach. Each data augmentation is conducted with 10\% probability. All experiments are conducted with an RTX 4090 and pytorch 2.0 version. 

\noindent\textbf{Evaluation metrics} We adopt detection accuracy and average precision in our experiments, which are widely used in previous studies. The number of real and fake images is balanced in all generative models.

\begin{figure*}[]
   \centering
   \includegraphics[width=6.8in,clip]{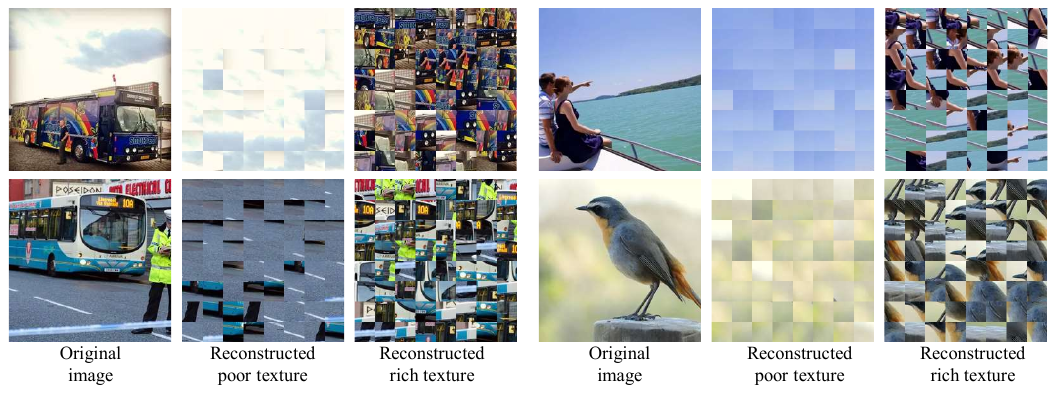}
   \caption{The illustration of \textit{Smash\&Reconstruction}.}
   \label{fig_smashrec}
\end{figure*}

\subsection{Detection Effectiveness}

Table \ref{tab_main} shows the detection accuracy comparison between our approach and baselines. Due to the space limitation, the average precision comparison can be found in the supplement.
Our approach outperforms baselines with a clear margin in the average detection accuracy. Although our approach only trained over ProGAN-based fake images, it still works effectively for a wide variety of diffusion-based fake images. Our approach can also identify fake images generated by impressive commercial APIs like Midjourney, whose algorithm is unknown. In terms of each generative model, our approach achieves satisfactory results in most cases. The performance only degrades for fake images generated by CycleGAN and GauGAN. The fake images generated by CycleGAN differ from those generated from random noise. The inputs of CycleGAN are real images, and the outputs are stylized images. For instance, CycleGAN transfers a horse into a zebra. The performance degradation of our detector may result from distribution discrepancy. DIRE-D performs effectively for most diffusion-based generative models. However, its accuracy drops significantly for GAN-based fake images. The reason is that the performance of DIRE depends on their assumption that diffusion-generated fake images are more likely to be reconstructed by a pre-trained diffusion model. The reconstruction error of real images is close to that of GAN-generated fake images. As a result, DIRE-D cannot identify GAN-based fake images. In addition, DIRE-D is trained with fake images generated by ADM, a diffusion-based generative model. It is not surprising that DIRE-D achieves high detection accuracy over diffusion-generated fake images. We also train DIRE detector from scratch over ProGAN-based fake images, which is the same as other baselines. DIRE-G can identify most GAN-generated fake images but fails to generalize to diffusion-generated fake images. Among all baselines, the average detection accuracy of LNP is closest to ours. It is still inferior to ours by 4\% over average detection accuracy.

In real-world applications, images spread on public platforms may undergo various common image processing like JPEG compression. Therefore, it is important to evaluate the performance of the detector handling with distorted images. We adopt three common image distortions, including JPEG compression (QF=95), Gaussian blur ($\sigma=1$), and image downsampling, where the image size is reduced to a quarter of its original size ($r=0.5$). Table \ref{tab_robust} shows the average detection accuracy of ours and baselines. The details of the detection accuracy of each generative model can be found in the supplement. Image distortion inevitably breaks and erases the artifacts left by generative models. As a result, the performance of all detectors (including ours) degrades for distorted images. However, our approach still achieves the best detection performance compared with others in most cases. These experiments also demonstrate that our fingerprint, inter-pixel correlation contrast between rich and poor texture regions, is more robust than baselines for distorted images. 

\begin{table}[]
  \centering
  
    \begin{tabular}{cccc}
    \toprule
          & \multicolumn{3}{c}{Distortion} \\
    Detector & JPEG  & Downsampling & Blur \\
    \hline
    CNNSpot & 64.03  & 58.85  & 68.39  \\
    FreDect & 66.95  & 35.84  & 65.75  \\
    Fusing & 62.43  & 50.00  & 68.09  \\
    GramNet & 65.47  & 60.30  & 68.63  \\
    LNP   & 53.56  & 63.28  & 65.88  \\
    LGrad & 51.55  & 60.86  & \underline{71.73}  \\
    DIRE-G & 66.49  & 56.09  & 64.00 \\
    DIRE-D & 70.27  & 62.26  & 70.46  \\
    UnivFD & \textbf{74.10}  & \underline{70.87}  & 70.31  \\
    Ours  & \underline{72.48}  & \textbf{78.36}  & \textbf{75.99}  \\
    \bottomrule
    \end{tabular}%
    \caption{Detection accuracy (average of various generative models) over distorted images. }
  \label{tab_robust}%
\end{table}%

\subsection{Ablation Studies}

\begin{table}[htbp]
  \centering
  
    \begin{tabular}{cccc}
    \toprule
    Generative model & w/o S\&R & w/o B. & w/o C. \\
    \hline
    ProGAN & 99.99  & 99.98  & 100.00  \\
    StyleGan & 92.10  & 89.33  & 94.31  \\
    BigGAN & 79.38  & 88.90  & 86.83  \\
    CycleGAN & 70.67  & 65.22  & 65.90  \\
    StarGAN & 89.29  & 82.92  & 99.22  \\
    GauGAN & 56.69  & 75.74  & 71.71  \\
    Stylegan2 & 90.23  & 92.78  & 95.18  \\
    whichfaceisreal & 58.65  & 81.25  & 76.55  \\
    ADM   & 62.88  & 64.99  & 77.00  \\
    Glide & 79.97  & 82.38  & 83.90  \\
    Midjourney & 57.23  & 90.37  & 82.53  \\
    SDv1.4 & 58.38  & 94.57  & 93.47  \\
    SDv1.5 & 58.78  & 94.69  & 93.22  \\
    VQDM  & 70.52  & 84.91  & 84.05  \\
    wukong & 56.65  & 89.29  & 86.00  \\
    DALLE2 & 91.95  & 96.75  & 95.20  \\
    SDXL  & 62.92  & 97.95  & 95.7  \\
    \hline
    Average & 73.34  & 85.88  &  87.10 \\
    Degradation & \textcolor{gray}{-17.13}  & \textcolor{gray}{-3.26}  & \textcolor{gray}{-2.74}  \\
    \bottomrule
    \end{tabular}%
    \caption{The impact of each component for detection accuracy. B. and C. denote boundary and contrast, respectively. The standard detection accuracy of our approach is 89.85. Each component is conducive to improving the detection accuracy.}
  \label{tab_fingerprint}%
\end{table}%

The crux of our approach is the fingerprint feature extraction, which mainly consists of \textit{Smash\&Reconstruction} and high-pass filters. Fig. \ref{fig_smashrec}  visualizes the results of \textit{Smash\&Reconstruction}. It aims to obtain the rich and poor texture regions from a spatial image. The semantic information of an image is almost destroyed. Erasing semantic information makes the detector NOT rely on the semantic information of images. Table \ref{tab_fingerprint} shows the evaluation results of our detector with spatial image inputs (high-pass filters are deployed). In this case named w/o S\&R, the performance of our detector significantly drops, especially for some diffusion-based models. The reason is that the detector for fake/real image authentication is over-reliance on the semantic information of the images. In our training set generated by ProGAN, some fake images are anomalous in the aspect of semantic information, which is easily perceived by humans. Therefore, our detector without \textit{Smash\&Reconstruction} distinguishes real or fake images based on whether their semantic information is reasonable. It fails to identify fake images whose semantic information is very close to real images, such as fake images generated by Midjourney or Stable Diffusion.

Since our rich/poor texture reconstructed image is made up of various patches, the boundary between different patches results in blocking artifact. We adopt high-pass filters across the boundaries to extract variation trends of patch diversity, which is conducive to the final detection accuracy. We try to remove blocking artifact by restricting the convolution range of high-pass filters. Specifically, high-pass filters do NOT cross the patch boundary and are only applied inside each patch. In this case named w/o Boundary, the detection accuracy of our detector slightly drops about 3.26\%. 

The contrast between rich and poor texture regions of fake images exhibits significant discrepancy from that of real images. Apart from spatial image input, we also evaluate the performance of our detector with patch-based reconstruction images. In this case named w/o Contrast, we randomly crop images into multiple patches and reconstruct images without depending on their texture diversity. The detector input is an image made up of randomly selected patches. Based on Table \ref{tab_fingerprint}, we find that the performance of this case is significantly better than that of spatial image input (w/o S\&R) but inferior to our approach. This result is also conformed to our intuition. The inter-pixel correlation contrast between rich and poor texture regions is effective in our approach.

More ablation studies including the patch size and the hige pass filters can be found in the supplement. 



\section{Conclusions}

In this paper, we propose a novel AI-generated image detection approach. The crux of the proposed approach is based on the inter-pixel correlation contrast between rich and poor texture regions. This fingerprint of fake images is universal across various generative models due to the inherent characteristic of distribution approximation. It is more difficult for generative models to synthesize rich texture regions from random noise due to the more complex distribution of rich texture regions compared with poor texture regions. This universal fingerprint of fake images boosts the generalization of our detector. Furthermore, we build a comprehensive AI-generated image detection benchmark, including various generative models and detectors. This benchmark facilitates the comparison for follow-up studies. Extensive experiments also demonstrate the superiority of the proposed approach compared with existing baselines.


{
    \small
    \bibliographystyle{ieeenat_fullname}
    \bibliography{main}

\begin{thebibliography}{53}
\providecommand{\natexlab}[1]{#1}
\providecommand{\url}[1]{\texttt{#1}}
\expandafter\ifx\csname urlstyle\endcsname\relax
  \providecommand{\doi}[1]{doi: #1}\else
  \providecommand{\doi}{doi: \begingroup \urlstyle{rm}\Url}\fi

\bibitem[Arik et~al.(2018)Arik, Chen, Peng, Ping, and Zhou]{arik2018neural}
Sercan Arik, Jitong Chen, Kainan Peng, Wei Ping, and Yanqi Zhou.
\newblock Neural voice cloning with a few samples.
\newblock \emph{Advances in neural information processing systems}, 31, 2018.

\bibitem[Barni et~al.(2020)Barni, Kallas, Nowroozi, and Tondi]{barni2020cnn}
Mauro Barni, Kassem Kallas, Ehsan Nowroozi, and Benedetta Tondi.
\newblock Cnn detection of gan-generated face images based on cross-band
  co-occurrences analysis.
\newblock In \emph{2020 IEEE international workshop on information forensics
  and security (WIFS)}, pages 1--6. IEEE, 2020.

\bibitem[Brock et~al.(2018)Brock, Donahue, and Simonyan]{brock2018large}
Andrew Brock, Jeff Donahue, and Karen Simonyan.
\newblock Large scale gan training for high fidelity natural image synthesis.
\newblock \emph{arXiv preprint arXiv:1809.11096}, 2018.

\bibitem[Cao et~al.(2022)Cao, Ma, Yao, Chen, Ding, and Yang]{cao2022end}
Junyi Cao, Chao Ma, Taiping Yao, Shen Chen, Shouhong Ding, and Xiaokang Yang.
\newblock End-to-end reconstruction-classification learning for face forgery
  detection.
\newblock In \emph{Proceedings of the IEEE/CVF Conference on Computer Vision
  and Pattern Recognition}, pages 4113--4122, 2022.

\bibitem[Chierchia et~al.(2014)Chierchia, Poggi, Sansone, and
  Verdoliva]{chierchia2014bayesian}
Giovanni Chierchia, Giovanni Poggi, Carlo Sansone, and Luisa Verdoliva.
\newblock A bayesian-mrf approach for prnu-based image forgery detection.
\newblock \emph{IEEE Transactions on Information Forensics and Security},
  9\penalty0 (4):\penalty0 554--567, 2014.

\bibitem[Choi et~al.(2018)Choi, Choi, Kim, Ha, Kim, and Choo]{choi2018stargan}
Yunjey Choi, Minje Choi, Munyoung Kim, Jung-Woo Ha, Sunghun Kim, and Jaegul
  Choo.
\newblock Stargan: Unified generative adversarial networks for multi-domain
  image-to-image translation.
\newblock In \emph{Proceedings of the IEEE conference on computer vision and
  pattern recognition}, pages 8789--8797, 2018.

\bibitem[cnn(2023)]{news}
cnn.
\newblock
  https://www.cnn.com/2023/05/22/tech/twitter-fake-image-pentagon-explosion/index.html.
\newblock 2023.

\bibitem[Deng et~al.(2009)Deng, Dong, Socher, Li, Li, and
  Fei-Fei]{deng2009imagenet}
Jia Deng, Wei Dong, Richard Socher, Li-Jia Li, Kai Li, and Li Fei-Fei.
\newblock Imagenet: A large-scale hierarchical image database.
\newblock In \emph{2009 IEEE conference on computer vision and pattern
  recognition}, pages 248--255. Ieee, 2009.

\bibitem[Dhariwal and Nichol(2021)]{dhariwal2021diffusion}
Prafulla Dhariwal and Alexander Nichol.
\newblock Diffusion models beat gans on image synthesis.
\newblock \emph{Advances in neural information processing systems},
  34:\penalty0 8780--8794, 2021.

\bibitem[Durall et~al.(2020)Durall, Keuper, and Keuper]{durall2020watch}
Ricard Durall, Margret Keuper, and Janis Keuper.
\newblock Watch your up-convolution: Cnn based generative deep neural networks
  are failing to reproduce spectral distributions.
\newblock In \emph{Proceedings of the IEEE/CVF conference on computer vision
  and pattern recognition}, pages 7890--7899, 2020.

\bibitem[Frank et~al.(2020)Frank, Eisenhofer, Sch{\"o}nherr, Fischer, Kolossa,
  and Holz]{frank2020leveraging}
Joel Frank, Thorsten Eisenhofer, Lea Sch{\"o}nherr, Asja Fischer, Dorothea
  Kolossa, and Thorsten Holz.
\newblock Leveraging frequency analysis for deep fake image recognition.
\newblock In \emph{International conference on machine learning}, pages
  3247--3258. PMLR, 2020.

\bibitem[Fridrich and Kodovsky(2012)]{fridrich2012rich}
Jessica Fridrich and Jan Kodovsky.
\newblock Rich models for steganalysis of digital images.
\newblock \emph{IEEE Transactions on information Forensics and Security},
  7\penalty0 (3):\penalty0 868--882, 2012.

\bibitem[Goodfellow et~al.(2020)Goodfellow, Pouget-Abadie, Mirza, Xu,
  Warde-Farley, Ozair, Courville, and Bengio]{goodfellow2020generative}
Ian Goodfellow, Jean Pouget-Abadie, Mehdi Mirza, Bing Xu, David Warde-Farley,
  Sherjil Ozair, Aaron Courville, and Yoshua Bengio.
\newblock Generative adversarial networks.
\newblock \emph{Communications of the ACM}, 63\penalty0 (11):\penalty0
  139--144, 2020.

\bibitem[Gragnaniello et~al.(2021)Gragnaniello, Cozzolino, Marra, Poggi, and
  Verdoliva]{gragnaniello2021gan}
Diego Gragnaniello, Davide Cozzolino, Francesco Marra, Giovanni Poggi, and
  Luisa Verdoliva.
\newblock Are gan generated images easy to detect? a critical analysis of the
  state-of-the-art.
\newblock In \emph{2021 IEEE international conference on multimedia and expo
  (ICME)}, pages 1--6. IEEE, 2021.

\bibitem[Gu et~al.(2022)Gu, Chen, Bao, Wen, Zhang, Chen, Yuan, and
  Guo]{gu2022vector}
Shuyang Gu, Dong Chen, Jianmin Bao, Fang Wen, Bo Zhang, Dongdong Chen, Lu Yuan,
  and Baining Guo.
\newblock Vector quantized diffusion model for text-to-image synthesis.
\newblock In \emph{Proceedings of the IEEE/CVF Conference on Computer Vision
  and Pattern Recognition}, pages 10696--10706, 2022.

\bibitem[Ho et~al.(2020)Ho, Jain, and Abbeel]{ho2020denoising}
Jonathan Ho, Ajay Jain, and Pieter Abbeel.
\newblock Denoising diffusion probabilistic models.
\newblock \emph{Advances in neural information processing systems},
  33:\penalty0 6840--6851, 2020.

\bibitem[Hulzebosch et~al.(2020)Hulzebosch, Ibrahimi, and
  Worring]{hulzebosch2020detecting}
Nils Hulzebosch, Sarah Ibrahimi, and Marcel Worring.
\newblock Detecting cnn-generated facial images in real-world scenarios.
\newblock In \emph{Proceedings of the IEEE/CVF conference on computer vision
  and pattern recognition workshops}, pages 642--643, 2020.

\bibitem[Ju et~al.(2022)Ju, Jia, Ke, Xue, Nagano, and Lyu]{ju2022fusing}
Yan Ju, Shan Jia, Lipeng Ke, Hongfei Xue, Koki Nagano, and Siwei Lyu.
\newblock Fusing global and local features for generalized ai-synthesized image
  detection.
\newblock In \emph{2022 IEEE International Conference on Image Processing
  (ICIP)}, pages 3465--3469. IEEE, 2022.

\bibitem[Karras et~al.(2017)Karras, Aila, Laine, and
  Lehtinen]{karras2017progressive}
Tero Karras, Timo Aila, Samuli Laine, and Jaakko Lehtinen.
\newblock Progressive growing of gans for improved quality, stability, and
  variation.
\newblock \emph{arXiv preprint arXiv:1710.10196}, 2017.

\bibitem[Karras et~al.(2019)Karras, Laine, and Aila]{karras2019style}
Tero Karras, Samuli Laine, and Timo Aila.
\newblock A style-based generator architecture for generative adversarial
  networks.
\newblock In \emph{Proceedings of the IEEE/CVF conference on computer vision
  and pattern recognition}, pages 4401--4410, 2019.

\bibitem[Karras et~al.(2020)Karras, Laine, Aittala, Hellsten, Lehtinen, and
  Aila]{karras2020analyzing}
Tero Karras, Samuli Laine, Miika Aittala, Janne Hellsten, Jaakko Lehtinen, and
  Timo Aila.
\newblock Analyzing and improving the image quality of stylegan.
\newblock In \emph{Proceedings of the IEEE/CVF conference on computer vision
  and pattern recognition}, pages 8110--8119, 2020.

\bibitem[Kim et~al.(2018)Kim, Garrido, Tewari, Xu, Thies, Niessner, P{\'e}rez,
  Richardt, Zollh{\"o}fer, and Theobalt]{kim2018deep}
Hyeongwoo Kim, Pablo Garrido, Ayush Tewari, Weipeng Xu, Justus Thies, Matthias
  Niessner, Patrick P{\'e}rez, Christian Richardt, Michael Zollh{\"o}fer, and
  Christian Theobalt.
\newblock Deep video portraits.
\newblock \emph{ACM transactions on graphics (TOG)}, 37\penalty0 (4):\penalty0
  1--14, 2018.

\bibitem[Kingma and Dhariwal(2018)]{kingma2018glow}
Durk~P Kingma and Prafulla Dhariwal.
\newblock Glow: Generative flow with invertible 1x1 convolutions.
\newblock \emph{Advances in neural information processing systems}, 31, 2018.

\bibitem[Kingma and Welling(2013)]{kingma2013auto}
Diederik~P Kingma and Max Welling.
\newblock Auto-encoding variational bayes.
\newblock \emph{arXiv preprint arXiv:1312.6114}, 2013.

\bibitem[Liu et~al.(2022)Liu, Yang, Bi, Xiao, Li, and Gao]{liu2022detecting}
Bo Liu, Fan Yang, Xiuli Bi, Bin Xiao, Weisheng Li, and Xinbo Gao.
\newblock Detecting generated images by real images.
\newblock In \emph{European Conference on Computer Vision}, pages 95--110.
  Springer, 2022.

\bibitem[Liu et~al.(2021)Liu, Li, Zhou, Chen, He, Xue, Zhang, and
  Yu]{liu2021spatial}
Honggu Liu, Xiaodan Li, Wenbo Zhou, Yuefeng Chen, Yuan He, Hui Xue, Weiming
  Zhang, and Nenghai Yu.
\newblock Spatial-phase shallow learning: rethinking face forgery detection in
  frequency domain.
\newblock In \emph{Proceedings of the IEEE/CVF conference on computer vision
  and pattern recognition}, pages 772--781, 2021.

\bibitem[Liu et~al.(2018)Liu, Luo, Wang, and Tang]{liu2018large}
Ziwei Liu, Ping Luo, Xiaogang Wang, and Xiaoou Tang.
\newblock Large-scale celebfaces attributes (celeba) dataset.
\newblock \emph{Retrieved August}, 15\penalty0 (2018):\penalty0 11, 2018.

\bibitem[Liu et~al.(2020)Liu, Qi, and Torr]{liu2020global}
Zhengzhe Liu, Xiaojuan Qi, and Philip~HS Torr.
\newblock Global texture enhancement for fake face detection in the wild.
\newblock In \emph{Proceedings of the IEEE/CVF conference on computer vision
  and pattern recognition}, pages 8060--8069, 2020.

\bibitem[Lukas et~al.(2006)Lukas, Fridrich, and Goljan]{lukas2006digital}
Jan Lukas, Jessica Fridrich, and Miroslav Goljan.
\newblock Digital camera identification from sensor pattern noise.
\newblock \emph{IEEE Transactions on Information Forensics and Security},
  1\penalty0 (2):\penalty0 205--214, 2006.

\bibitem[Midjourney(2023)]{midjourney}
Midjourney.
\newblock https://www.midjourney.com/home/.
\newblock 2023.

\bibitem[Nichol et~al.(2021)Nichol, Dhariwal, Ramesh, Shyam, Mishkin, McGrew,
  Sutskever, and Chen]{nichol2021glide}
Alex Nichol, Prafulla Dhariwal, Aditya Ramesh, Pranav Shyam, Pamela Mishkin,
  Bob McGrew, Ilya Sutskever, and Mark Chen.
\newblock Glide: Towards photorealistic image generation and editing with
  text-guided diffusion models.
\newblock \emph{arXiv preprint arXiv:2112.10741}, 2021.

\bibitem[Ojha et~al.(2023)Ojha, Li, and Lee]{ojha2023towards}
Utkarsh Ojha, Yuheng Li, and Yong~Jae Lee.
\newblock Towards universal fake image detectors that generalize across
  generative models.
\newblock In \emph{Proceedings of the IEEE/CVF Conference on Computer Vision
  and Pattern Recognition}, pages 24480--24489, 2023.

\bibitem[Park et~al.(2019)Park, Liu, Wang, and Zhu]{park2019semantic}
Taesung Park, Ming-Yu Liu, Ting-Chun Wang, and Jun-Yan Zhu.
\newblock Semantic image synthesis with spatially-adaptive normalization.
\newblock In \emph{Proceedings of the IEEE/CVF conference on computer vision
  and pattern recognition}, pages 2337--2346, 2019.

\bibitem[Pu et~al.(2022)Pu, Hu, Wang, Li, Hu, Zhu, Song, Song, Wu, and
  Lyu]{pu2022learning}
Wenbo Pu, Jing Hu, Xin Wang, Yuezun Li, Shu Hu, Bin Zhu, Rui Song, Qi Song, Xi
  Wu, and Siwei Lyu.
\newblock Learning a deep dual-level network for robust deepfake detection.
\newblock \emph{Pattern Recognition}, 130:\penalty0 108832, 2022.

\bibitem[Qiao et~al.(2019)Qiao, Zhang, Xu, and Tao]{qiao2019mirrorgan}
Tingting Qiao, Jing Zhang, Duanqing Xu, and Dacheng Tao.
\newblock Mirrorgan: Learning text-to-image generation by redescription.
\newblock In \emph{Proceedings of the IEEE/CVF Conference on Computer Vision
  and Pattern Recognition}, pages 1505--1514, 2019.

\bibitem[Radford et~al.(2021)Radford, Kim, Hallacy, Ramesh, Goh, Agarwal,
  Sastry, Askell, Mishkin, Clark, et~al.]{radford2021learning}
Alec Radford, Jong~Wook Kim, Chris Hallacy, Aditya Ramesh, Gabriel Goh,
  Sandhini Agarwal, Girish Sastry, Amanda Askell, Pamela Mishkin, Jack Clark,
  et~al.
\newblock Learning transferable visual models from natural language
  supervision.
\newblock In \emph{International conference on machine learning}, pages
  8748--8763. PMLR, 2021.

\bibitem[Ramesh et~al.(2022)Ramesh, Dhariwal, Nichol, Chu, and Chen]{dall-e-2}
Aditya Ramesh, Prafulla Dhariwal, Alex Nichol, Casey Chu, and Mark Chen.
\newblock Hierarchical text-conditional image generation with clip latents.
\newblock \emph{arXiv preprint arXiv:2204.06125}, 1\penalty0 (2):\penalty0 3,
  2022.

\bibitem[Reed et~al.(2016)Reed, Akata, Yan, Logeswaran, Schiele, and
  Lee]{reed2016generative}
Scott Reed, Zeynep Akata, Xinchen Yan, Lajanugen Logeswaran, Bernt Schiele, and
  Honglak Lee.
\newblock Generative adversarial text to image synthesis.
\newblock In \emph{International conference on machine learning}, pages
  1060--1069. PMLR, 2016.

\bibitem[Rombach et~al.(2022)Rombach, Blattmann, Lorenz, Esser, and
  Ommer]{rombach2022high}
Robin Rombach, Andreas Blattmann, Dominik Lorenz, Patrick Esser, and Bj{\"o}rn
  Ommer.
\newblock High-resolution image synthesis with latent diffusion models.
\newblock In \emph{Proceedings of the IEEE/CVF conference on computer vision
  and pattern recognition}, pages 10684--10695, 2022.

\bibitem[Rossler et~al.(2019)Rossler, Cozzolino, Verdoliva, Riess, Thies, and
  Nie{\ss}ner]{rossler2019faceforensics++}
Andreas Rossler, Davide Cozzolino, Luisa Verdoliva, Christian Riess, Justus
  Thies, and Matthias Nie{\ss}ner.
\newblock Faceforensics++: Learning to detect manipulated facial images.
\newblock In \emph{Proceedings of the IEEE/CVF international conference on
  computer vision}, pages 1--11, 2019.

\bibitem[Scherhag et~al.(2019)Scherhag, Debiasi, Rathgeb, Busch, and
  Uhl]{scherhag2019detection}
Ulrich Scherhag, Luca Debiasi, Christian Rathgeb, Christoph Busch, and Andreas
  Uhl.
\newblock Detection of face morphing attacks based on prnu analysis.
\newblock \emph{IEEE Transactions on Biometrics, Behavior, and Identity
  Science}, 1\penalty0 (4):\penalty0 302--317, 2019.

\bibitem[Sha et~al.(2022)Sha, Li, Yu, and Zhang]{sha2022fake}
Zeyang Sha, Zheng Li, Ning Yu, and Yang Zhang.
\newblock De-fake: Detection and attribution of fake images generated by
  text-to-image diffusion models.
\newblock \emph{arXiv preprint arXiv:2210.06998}, 2022.

\bibitem[Sohl-Dickstein et~al.(2015)Sohl-Dickstein, Weiss, Maheswaranathan, and
  Ganguli]{sohl2015deep}
Jascha Sohl-Dickstein, Eric Weiss, Niru Maheswaranathan, and Surya Ganguli.
\newblock Deep unsupervised learning using nonequilibrium thermodynamics.
\newblock In \emph{International conference on machine learning}, pages
  2256--2265. PMLR, 2015.

\bibitem[Song and Ermon(2019)]{song2019generative}
Yang Song and Stefano Ermon.
\newblock Generative modeling by estimating gradients of the data distribution.
\newblock \emph{Advances in neural information processing systems}, 32, 2019.

\bibitem[Suwajanakorn et~al.(2017)Suwajanakorn, Seitz, and
  Kemelmacher-Shlizerman]{suwajanakorn2017synthesizing}
Supasorn Suwajanakorn, Steven~M Seitz, and Ira Kemelmacher-Shlizerman.
\newblock Synthesizing obama: learning lip sync from audio.
\newblock \emph{ACM Transactions on Graphics (ToG)}, 36\penalty0 (4):\penalty0
  1--13, 2017.

\bibitem[Tan et~al.(2023)Tan, Zhao, Wei, Gu, and Wei]{tan2023learning}
Chuangchuang Tan, Yao Zhao, Shikui Wei, Guanghua Gu, and Yunchao Wei.
\newblock Learning on gradients: Generalized artifacts representation for
  gan-generated images detection.
\newblock In \emph{Proceedings of the IEEE/CVF Conference on Computer Vision
  and Pattern Recognition}, pages 12105--12114, 2023.

\bibitem[Wang et~al.(2020)Wang, Wang, Zhang, Owens, and Efros]{wang2020cnn}
Sheng-Yu Wang, Oliver Wang, Richard Zhang, Andrew Owens, and Alexei~A Efros.
\newblock Cnn-generated images are surprisingly easy to spot... for now.
\newblock In \emph{Proceedings of the IEEE/CVF conference on computer vision
  and pattern recognition}, pages 8695--8704, 2020.

\bibitem[Wang et~al.(2023)Wang, Bao, Zhou, Wang, Hu, Chen, and
  Li]{wang2023dire}
Zhendong Wang, Jianmin Bao, Wengang Zhou, Weilun Wang, Hezhen Hu, Hong Chen,
  and Houqiang Li.
\newblock Dire for diffusion-generated image detection.
\newblock \emph{arXiv preprint arXiv:2303.09295}, 2023.

\bibitem[West and Bergstrom(2023)]{whichfaceisreal}
Jevin West and Carl Bergstrom.
\newblock https://www.whichfaceisreal.com/.
\newblock 2023.

\bibitem[wukong(2023)]{wukong}
wukong.
\newblock https://xihe.mindspore.cn/modelzoo/wukong.
\newblock 2023.

\bibitem[Yu et~al.(2015)Yu, Seff, Zhang, Song, Funkhouser, and
  Xiao]{yu2015lsun}
Fisher Yu, Ari Seff, Yinda Zhang, Shuran Song, Thomas Funkhouser, and Jianxiong
  Xiao.
\newblock Lsun: Construction of a large-scale image dataset using deep learning
  with humans in the loop.
\newblock \emph{arXiv preprint arXiv:1506.03365}, 2015.

\bibitem[Zhu et~al.(2017)Zhu, Park, Isola, and Efros]{zhu2017unpaired}
Jun-Yan Zhu, Taesung Park, Phillip Isola, and Alexei~A Efros.
\newblock Unpaired image-to-image translation using cycle-consistent
  adversarial networks.
\newblock In \emph{Proceedings of the IEEE international conference on computer
  vision}, pages 2223--2232, 2017.

\bibitem[Zhu et~al.(2023)Zhu, Chen, Yan, Huang, Lin, Li, Tu, Hu, Hu, and
  Wang]{zhu2023genimage}
Mingjian Zhu, Hanting Chen, Qiangyu Yan, Xudong Huang, Guanyu Lin, Wei Li,
  Zhijun Tu, Hailin Hu, Jie Hu, and Yunhe Wang.
\newblock Genimage: A million-scale benchmark for detecting ai-generated image.
\newblock \emph{arXiv preprint arXiv:2306.08571}, 2023.

\end{thebibliography}
}

\clearpage

\maketitle

\section*{Supplement}

Our supplement consists of five parts, i.e., the parameters of high-pass filters, the structure of the classifier, the details experimental results of detection performance, complementary ablation studies, and AI-generated fake image visualization.

\begin{figure}[ht]
	\centering
  \includegraphics[width=3.2in,clip,trim=70 165 610 70]{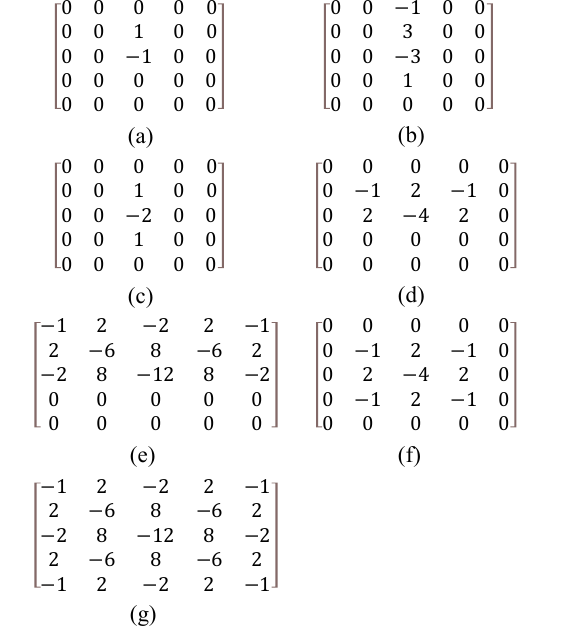}
	\caption{The specific kernel parameters of high-pass filters. }\label{fig_hpf}
\end{figure}

\subsection*{High-pass Filters}
In our manuscript, we adopt 30 high-pass filters which are proposed in the image steganalysis domain. Various filters can project spatial images in different directions and extract diverse high frequency signals. Fig. \ref{fig_hpf} shows the specific parameters of high-pass filters. We show 7 kernel parameters for concision. The entire kernel parameters are as follows. Eight different variants of (a) by rotating (a) following eight directions \{ $\nearrow, \rightarrow, \searrow, \downarrow, \swarrow \leftarrow, \nwarrow, \uparrow $ \}. (b) is similar to (a). Four different variants of (c) by rotating (c) following four directions \{ $ \rightarrow,  \downarrow, \nearrow ,\searrow  $ \} (\{ $ \leftarrow ,\uparrow ,\swarrow ,\nwarrow  $ \} are the same as their reverse directions therefore we ignore them.). Four different variants of (d) by rotating (d) following four directions \{ $ \rightarrow,  \downarrow, \leftarrow, \uparrow $ \}. (e) is similar to (d). Therefore, we obtain 2*8+1*4+2*4+2=30 high-pass filters.

\subsection*{Classifier Structure}

Since our proposed fingerprint exists across various AI-generated fake images, a simple cascaded CNN network can achieve satisfactory detection performance. The structure of our classifier is shown in Table \ref{tab_classifier}.

\begin{table}[htbp]
  \centering
    \begin{tabular}{cccc}
    \toprule
    Type  & Kernel num & With BN & Activation \\
    \hline
    Convo. & 32    & TRUE  & ReLU \\
    Convo. & 32    & TRUE  & ReLU \\
    Convo. & 32    & TRUE  & ReLU \\
    Convo. & 32    & TRUE  & ReLU \\
    Avg Pooling & None  & None  & None \\
    Convo. & 32    & TRUE  & ReLU \\
    Convo. & 32    & TRUE  & ReLU \\
    Avg Pooling & None  & None  & None \\
    Convo. & 32    & TRUE  & ReLU \\
    Convo. & 32    & TRUE  & ReLU \\
    Avg Pooling & None  & None  & None \\
    Convo. & 32    & TRUE  & ReLU \\
    Convo. & 32    & TRUE  & ReLU \\
    AdpAvgPool & None  & None  & None \\
    Flatten & None  & None  & None \\
    FC    & None  & FALSE & None \\
    \bottomrule
    \end{tabular}%
    \caption{The structure of the classifier.}
  \label{tab_classifier}%
\end{table}%

\subsection*{Performance Evaluation}
In the manuscript, we adopt accuracy and average precision as the criterion for AI-generated fake image detection. Table \ref{tab_acc_nodistortion} -\ref{tab_map_Blur} shows the concrete experimental results. 

\subsection*{Complementary Ablation Studies}

\begin{table}[htbp]
  \centering
  
    \begin{tabular}{ccccc}
    \toprule
          & \multirow{2}[0]{*}{w/o HPF} & \multicolumn{2}{c}{Patch Size} & \multirow{2}[0]{*}{Ours} \\
    Generator &       & 16    & 64    &  \\
    \hline
    ProGAN & 100.00  & 100.00  & 99.99  & 100.00  \\
    StyleGAN & 88.35  & 92.82  & 92.56  & 92.77  \\
    BigGAN & 79.35  & 84.78  & 94.25  & 95.80  \\
    CycleGAN & 69.57  & 54.16  & 68.02  & 70.17  \\
    StarGAN & 78.21  & 77.91  & 99.20  & 99.97  \\
    GauGAN & 64.26  & 50.71  & 68.50  & 71.58  \\
    Stylegan2 & 93.41  & 92.44  & 89.43  & 89.55  \\
    whichfaceisreal & 65.85  & 82.70  & 89.80  & 85.80  \\
    ADM   & 63.69  & 57.93  & 72.42  & 82.17  \\
    Glide & 71.03  & 84.99  & 79.77  & 83.79  \\
    Midjourney & 72.26  & 83.40  & 90.38  & 90.12  \\
    SDv1.4 & 87.35  & 92.78  & 95.71  & 95.38  \\
    SDv1.5 & 87.24  & 92.88  & 95.91  & 95.30  \\
    VQDM  & 69.71  & 75.76  & 89.83  & 88.91  \\
    Wukong & 78.82  & 86.90  & 89.88  & 91.07  \\
    DALLE2 & 80.85  & 96.35  & 93.80  & 96.60  \\
    SDXL & 84.97  & 97.72  & 98.58  & 98.43  \\
    \hline
    Average & 78.12  & 81.66  & 88.09  & 89.85  \\
    Degradation & \textcolor{gray}{-11.32}  & \textcolor{gray}{-7.25}  & \textcolor{gray}{-1.18}  \\
    \bottomrule
    \end{tabular}%
    \caption{The ablation studies of high-pass filters and the patch size.}
  \label{tab_ablation}%
\end{table}%

The size of rich/poor texture reconstructed images is based on the patch size and the number of patches. In the above experiments, we set the size of rich/poor texture region image as $256\times256$, and the size of each patch is $32\times 32$. In other words, each texture region image is made up of 64 patches. In the ablation experiments, we analyze the impact of patch size on the final performance. We set the size of the rich/poor patch as $16\times 16$ or $64\times 64$. The number of the reconstructed image is 256 or 16, respectively. The experimental results of the different patch sizes are shown in Table \ref{tab_ablation}. Our approach achieves the best performance with $32\times32$ patch size. The detection performance drops with a small patch size.

An alternative important component of fingerprint feature extraction is high-pass filters. Previous studies show that generated images exhibit more artifacts in the high frequency domain. Therefore, we adopt high-pass filters to magnify fake image artifacts and make the detector focus on the inter-pixel correlations. Based on Table \ref{tab_ablation} (w/o HPF), high-pass filters can improve our detection accuracy by 11\% for average cases.

\subsection*{AI-generated fake image visualization}

\begin{figure*}[ht]
	\centering
  \includegraphics[width=6.8in,clip,trim=250 390 360 0]{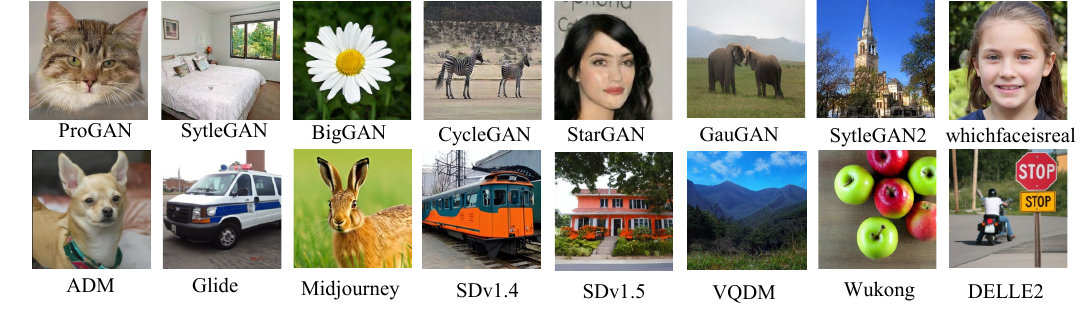}
	\caption{The visualization results of AI-generated fake images.}
\end{figure*}

In this supplement, we visualize various generative models including GAN-based models, Diffusion-based models, and their variants. Modern diffusion-based generator like Midjourney can create realistic-look synthetic images, which is hardly distinguished from real ones by humans.

\begin{table*}[]
  \centering
  
  \small
    \begin{tabular}{ccccccccccc}
    \toprule
    Generator & CNNSpot & FreDect & Fusing & GramNet & LNP   & LGrad & DIRE-G & DIRE-D & UnivFD & Ours \\
    \hline
    ProGAN & \textbf{100.00} & 99.36 & \textbf{100.00} & 99.99 & 99.95 & 99.83 & 95.19 & 52.75 & 99.81 & \textbf{100.00} \\
    StyleGan & 90.17 & 78.02 & 85.20 & 87.05 & \underline{ 92.64} & 91.08 & 83.03 & 51.31 & 84.93 & \textbf{92.77} \\
    BigGAN & 71.17 & 81.97 & 77.40 & 67.33 & 88.43 & 85.62 & 70.12 & 49.70 & \underline{ 95.08} & \textbf{95.80} \\
    CycleGAN & \underline{ 87.62} & 78.77 & 87.00 & 86.07 & 79.07 & 86.94 & 74.19 & 49.58 & \textbf{98.33} & 70.17 \\
    StarGAN & 94.60 & 94.62 & 97.00 & 95.05 & \textbf{100.00} & 99.27 & 95.47 & 46.72 & 95.75 & \underline{99.97} \\
    GauGAN & \underline{ 81.42} & 80.57 & 77.00 & 69.35 & 79.17 & 78.46 & 67.79 & 51.23 & \textbf{99.47} & 71.58 \\
    Stylegan2 & 86.91 & 66.19 & 83.30 & 87.28 & \textbf{93.82} & 85.32 & 75.31 & 51.72 & 74.96 & \underline{ 89.55} \\
    whichfaceisreal & \textbf{91.65} & 50.75 & 66.80 & 86.80 & 50.00 & 55.70 & 58.05 & 53.30 & \underline{ 86.90} & 85.80 \\
    ADM & 60.39 & 63.42 & 49.00 & 58.61 & \underline{ 83.91} & 67.15 & 75.78 & \textbf{98.25} & 66.87 & 82.17 \\
    Glide & 58.07 & 54.13 & 57.20 & 54.50 & 83.50 & 66.11 & 71.75 & \textbf{92.42} & 62.46 & \underline{ 83.79} \\
    Midjourney & 51.39 & 45.87 & 52.20 & 50.02 & 69.55 & 65.35 & 58.01 & \underline{ 89.45} & 56.13 & \textbf{90.12} \\
    SDv1.4 & 50.57 & 38.79 & 51.00 & 51.70 & 89.33 & 63.02 & 49.74 & \underline{ 91.24} & 63.66 & \textbf{95.38} \\
    SDv1.5 & 50.53 & 39.21 & 51.40 & 52.16 & 88.81 & 63.67 & 49.83 & \underline{ 91.63} & 63.49 & \textbf{95.30} \\
    VQDM & 56.46 & 77.80 & 55.10 & 52.86 & 85.03 & 72.99 & 53.68 & \textbf{91.90} & 85.31 & \underline{ 88.91} \\
    wukong & 51.03 & 40.30 & 51.70 & 50.76 & 86.39 & 59.55 & 54.46 & \underline{ 90.90} & 70.93 & \textbf{91.07} \\
    DALLE2 & 50.45 & 34.70 & 52.80 & 49.25 & 92.45 & 65.45 & 66.48 & \underline{ 92.45} & 50.75 & \textbf{96.60} \\
    SDXL & 53.03 & 51.23 & 55.60 & 64.53 & 87.75 & 71.30 & 55.35 & \underline{ 91.28} & 50.73 & \textbf{98.43} \\
    \hline
    Average & 69.73 & 63.28 & 67.63 & 68.43 & \underline{ 85.28} & 75.11 & 67.90 & 72.70 & 76.80 & \textbf{89.85}\\
    \bottomrule
    \end{tabular}%
    \caption{The detection accuracy (no distortion) comparison between our approach and baselines.}
  \label{tab_acc_nodistortion}%
\end{table*}%

\begin{table*}[]
  \centering
  
  \small
    \begin{tabular}{ccccccccccc}
    \toprule
    Generative   model & CNNSpot & FreDect & Fusing & GramNet & LNP & LGrad & DIRE-G & DIRE-D & UnivFD & Ours \\
    \hline
    ProGAN & \textbf{100.00} & \underline{ 99.99} & \textbf{100.00} & \textbf{100.00} & \textbf{100.00} & \textbf{100.00} & 99.08 & 58.79 & \textbf{100.00} & \textbf{100.00} \\
    StyleGan & \textbf{99.83} & 88.98 & \underline{ 99.50} & 99.23 & 99.27 & 98.31 & 91.74 & 56.68 & 97.56 & 98.96 \\
    BigGAN & 85.99 & 93.62 & 90.70 & 81.79 & 94.54 & 92.93 & 75.25 & 46.91 & \underline{ 99.27} & \textbf{99.42} \\
    CycleGAN & 94.94 & 84.78 & \underline{ 95.50} & 95.33 & 89.52 & 95.01 & 80.56 & 50.03 & \textbf{99.80} & 85.26 \\
    StarGAN & 99.04 & 99.49 & \underline{ 99.80} & 99.23 & \textbf{100.00} & \textbf{100.00} & 99.34 & 40.64 & 99.37 & \textbf{100.00} \\
    GauGAN & 90.82 & 82.84 & 88.30 & 84.99 & 84.54 & \underline{ 95.43} & 72.15 & 47.34 & \textbf{99.98} & 81.33 \\
    Stylegan2 & 99.48 & 82.54 & \underline{ 99.60} & 99.11 & \textbf{99.70} & 97.89 & 88.29 & 58.03 & 97.90 & 97.74 \\
    whichfaceisreal & \textbf{99.85} & 55.85 & 93.30 & 95.21 & 42.75 & 57.99 & 60.13 & 59.02 & \underline{ 96.73} & 95.26 \\
    ADM & 75.67 & 61.77 & \underline{ 94.10} & 73.11 & 93.37 & 72.95 & 85.84 & \textbf{99.79} & 86.81 & 93.40 \\
    Glide & 72.28 & 52.92 & 77.50 & 66.76 & 92.76 & 80.42 & 78.35 & \textbf{99.54} & 83.81 & \underline{ 94.04} \\
    Midjourney & 66.24 & 46.09 & 70.00 & 56.82 & 86.92 & 71.86 & 61.86 & \textbf{97.32} & 74.00 & \underline{ 96.48} \\
    SDv1.4 & 61.20 & 37.83 & 65.40 & 59.83 & 96.34 & 62.37 & 49.87 & \underline{ 98.61} & 86.14 & \textbf{99.06} \\
    SDv1.5 & 61.56 & 37.76 & 65.70 & 60.37 & 96.00 & 62.85 & 49.52 & \underline{ 98.83} & 85.84 & \textbf{99.06} \\
    VQDM & 68.83 & 85.10 & 75.60 & 61.13 & 94.91 & 77.47 & 54.57 & \textbf{98.98} & \underline{ 96.53} & 96.26 \\
    wukong & 57.34 & 39.58 & 64.60 & 55.62 & 95.33 & 62.48 & 55.38 & \textbf{98.37} & 91.07 & \underline{ 97.54} \\
    DALLE2 & 53.51 & 38.20 & 68.12 & 49.82 & 98.26 & 82.55 & 74.48 & \textbf{99.71} & 63.04 & \underline{ 99.56} \\
    SDXL & 72.62 & 49.45 & 79.41 & 68.22 & 87.75 & 80.03 & 53.97 & \underline{ 99.10} & 67.59 & \textbf{99.89} \\
    \hline
    Average & 79.95 & 66.87 & 83.95 & 76.86 & \underline{ 91.29} & 81.80 & 72.38 & 76.92 & 89.73 & \textbf{96.07} \\
    \bottomrule
    \end{tabular}%
    \caption{The average precision (no distortion) comparison between our approach and baselines.}
  \label{tab_map_nodistortion}%
\end{table*}%

\begin{table*}[]
  \centering
  
  \small
    \begin{tabular}{ccccccccccc}
    \toprule
        Generator & CNNSpot & FreDect & Fusing & GramNet & LNP   & LGrad & DIRE-G & DIRE-D & UnivFD & Ours \\
        \hline
    ProGAN & \textbf{99.96} & 84.40 & 99.64 & \underline{ 99.94} & 67.80 & 55.70 & 98.92 & 51.55 & 99.34 & 97.84 \\
    StyleGan & 75.00 & 72.30 & 70.45 & 77.74 & 55.42 & 55.90 & 75.93 & 50.89 & \underline{ 79.74} & \textbf{82.49} \\
    BigGAN & 62.05 & 62.90 & 63.50 & 62.28 & 51.73 & 51.35 & \underline{ 67.85} & 51.08 & \textbf{88.22} & 65.25 \\
    CycleGAN & 83.16 & 71.35 & 80.96 & \underline{ 86.98} & 62.34 & 57.15 & 74.68 & 50.53 & \textbf{96.14} & 71.99 \\
    StarGAN & 79.26 & 83.79 & \underline{ 92.02} & 88.72 & 52.00 & 51.68 & 83.08 & 41.55 & \textbf{95.32} & 60.21 \\
    GauGAN & 69.89 & 64.99 & 66.91 & 65.69 & 50.18 & 49.70 & \underline{ 77.84} & 39.31 & \textbf{98.61} & 73.71 \\
    Stylegan2 & 71.29 & 73.34 & 60.19 & \underline{ 78.87} & 58.32 & 55.99 & 68.99 & 51.36 & 69.41 & \textbf{82.71} \\
    whichfaceisreal & 79.10 & 50.25 & 59.90 & \textbf{86.40} & 50.13 & 52.90 & 59.05 & 54.35 & 70.20 & \underline{ 79.40} \\
    ADM & 51.28 & \underline{ 67.72} & 50.01 & 51.62 & 51.22 & 46.50 & 62.49 & \textbf{91.92} & 64.41 & 62.64 \\
    Glide & 51.92 & 66.13 & 51.99 & 51.59 & 51.28 & 45.09 & \underline{ 73.11} & \textbf{91.38} & 64.14 & 68.01 \\
    Midjourney & 50.90 & 56.61 & 50.85 & 49.85 & 50.14 & \underline{ 62.32} & 58.90 & \textbf{90.78} & 55.43 & 57.87 \\
    SDv1.4 & 49.82 & 52.95 & 50.16 & 49.44 & 51.84 & 53.87 & 48.40 & \textbf{92.53} & 56.48 & \underline{ 75.00} \\
    SDv1.5 & 49.90 & 52.71 & 50.13 & 49.56 & 51.87 & 53.62 & 48.48 & \textbf{92.56} & 56.19 & \underline{ 74.87} \\
    VQDM & 51.24 & 76.51 & 51.61 & 50.98 & 51.06 & 44.06 & 54.82 & \textbf{92.47} & \underline{ 78.89} & 64.94 \\
    wukong & 50.01 & 52.60 & 50.50 & 49.62 & 50.91 & 52.49 & 50.88 & \textbf{92.14} & 62.75 & \underline{ 67.91} \\
    DALLE2 & 49.70 & \underline{ 82.70} & 50.00 & 48.20 & 50.70 & 36.40 & 60.40 & \textbf{89.89} & 50.35 & 70.35 \\
    SDXL & 49.98 & \underline{ 75.00} & 50.48 & 49.93 & 49.93 & 65.98 & 56.53 & \textbf{91.13} & 50.65 & 69.00 \\
    \hline
    Average & 63.20 & 67.43 & 61.72 & 64.55 & 53.35 & 52.39 & 65.90 & 71.50 &  \textbf{72.72} & \underline{ 72.01} \\
    \bottomrule
    \end{tabular}%
    \caption{The detection accuracy (JPG compression) comparison between our approach and baselines.}
  \label{tab_acc_jpg}%
\end{table*}%

\begin{table*}[]
  \centering
  
  \small
    \begin{tabular}{ccccccccccc}
    \toprule
    Generative   model & CNNSpot & FreDect & Fusing & GramNet & LNP & LGrad & DIRE-G & DIRE-D & UnivFD & Ours \\
    \hline
    ProGAN & \textbf{100.00} & 95.45 & \textbf{100.00} & \textbf{100.00} & 91.04 & 68.65 & 99.94 & 50.95 & \underline{ 99.98} & 99.73 \\
    StyleGan & \underline{ 98.60} & 79.65 & \textbf{98.62} & 98.24 & 79.86 & 75.48 & 94.12 & 53.59 & 94.90 & 91.12 \\
    BigGAN & 88.36 & 69.81 & \underline{ 91.39} & 83.14 & 67.56 & 54.21 & 78.07 & 50.21 & \textbf{97.65} & 69.41 \\
    CycleGAN & 96.08 & 82.42 & \underline{ 97.13} & 96.32 & 90.39 & 70.26 & 84.29 & 52.13 & \textbf{99.47} & 83.3 \\
    StarGAN & 93.63 & 94.37 & \textbf{99.29} & 97.16 & 76.68 & 61.44 & 92.16 & 39.44 & \underline{ 99.07} & 70.29 \\
    GauGAN & 95.53 & 68.76 & \underline{ 96.69} & 89.54 & 60.98 & 51.42 & 76.22 & 37.73 & \textbf{99.91} & 82.41 \\
    Stylegan2 & 98.08 & 82.48 & \underline{ 98.26} & \textbf{98.42} & 87.56 & 77.57 & 92.02 & 54.50 & 94.65 & 91.03 \\
    whichfaceisreal & 94.34 & 54.14 & \textbf{95.27} & \underline{ 95.19} & 69.81 & 55.02 & 61.86 & 60.36 & 88.13 & 89.49 \\
    ADM & 64.18 & 87.14 & \underline{ 88.56} & 60.64 & 61.90 & 44.73 & 71.20 & \textbf{98.99} & 84.00 & 75.07 \\
    Glide & 69.60 & \underline{ 91.06} & 80.01 & 62.67 & 69.27 & 42.16 & 83.60 & \textbf{98.44} & 85.18 & 81.04 \\
    Midjourney & 61.95 & 72.61 & \underline{ 75.21} & 54.79 & 63.94 & 68.04 & 65.10 & \textbf{97.96} & 71.41 & 66.42 \\
    SDv1.4 & 55.52 & 66.99 & 64.10 & 52.56 & 70.53 & 56.03 & 45.14 & \textbf{99.15} & 75.23 & \underline{ 86.47} \\
    SDv1.5 & 56.47 & 66.15 & 64.28 & 53.32 & 70.51 & 56.30 & 45.22 & \textbf{99.19} & 74.82 & \underline{ 86.59} \\
    VQDM & 64.82 & 91.26 & 76.46 & 56.98 & 60.51 & 40.91 & 58.94 & \textbf{99.39} & \underline{ 93.58} & 78.54 \\
    wukong & 54.78 & 65.54 & 63.78 & 51.40 & 66.00 & 54.32 & 50.50 & \textbf{99.23} & \underline{ 83.69} & 80.26 \\
    DALLE2 & 47.03 & \underline{ 98.58} & 54.97 & 43.52 & 55.31 & 36.37 & 66.37 & \textbf{99.51} & 58.58 & 95.59 \\
    SDXL & 75.26 & \underline{ 94.45} & 80.73 & 43.63 & 50.82 & 73.81 & 60.10 & \textbf{99.15} & 64.97 & 86.12 \\
    \hline
    Average & 77.31 & 80.05 & \underline{ 83.81} & 72.80 & 70.16 & 58.04 & 72.05 & 75.88 & \textbf{86.19} & 83.11 \\
        \bottomrule
    \end{tabular}
    \caption{The average precision (JPG compression) comparison between our approach and baselines.}
  \label{tab_map_jpg}%
\end{table*}%

\begin{table*}[]
  \centering
  
  \small
    \begin{tabular}{ccccccccccc}
    \toprule
        Generator & CNNSpot & FreDect & Fusing & GramNet & LNP   & LGrad & DIRE-G & DIRE-D & UnivFD & Ours \\
        \hline
    ProGAN & 88.00 & 60.85 & 50.00 & 88.30 & 85.15 & 81.46 & 68.01 & 49.74 & \underline{ 95.83} & \textbf{99.92} \\
    StyleGan & 64.46 & 61.81 & 50.00 & 67.69 & 80.34 & 71.32 & 66.52 & 51.74 & \underline{ 72.63} & \textbf{90.37} \\
    BigGAN & 52.02 & 52.70 & 50.00 & 52.20 & 70.95 & 58.23 & 52.08 & 50.68 & \textbf{73.00} & \underline{ 72.35} \\
    CycleGAN & 60.83 & 48.75 & 50.00 & 67.41 & 67.79 & 52.42 & 57.53 & 46.21 & \textbf{89.21} & \underline{ 83.76} \\
    StarGAN & 64.58 & 51.68 & 50.00 & 77.89 & 86.44 & 58.38 & 59.80 & 58.39 & \underline{ 88.02} & \textbf{99.90} \\
    GauGAN & \underline{ 65.61} & 50.55 & 50.00 & 63.97 & 53.89 & 55.04 & 44.82 & 52.61 & \textbf{91.23} & 62.07 \\
    Stylegan2 & 65.69 & 63.50 & 50.00 & 69.40 & \textbf{89.16} & 69.93 & 66.79 & 60.05 & 68.66 & \underline{ 89.00} \\
    whichfaceisreal & 76.50 & 49.70 & 50.00 & \textbf{79.80} & 51.90 & 56.70 & 50.85 & 50.00 & 73.75 & \underline{ 79.55} \\
    ADM & 50.09 & 15.68 & 49.98 & 49.38 & 66.25 & 55.89 & 54.63 & \textbf{75.27} & \underline{ 71.94} & 71.12 \\
    Glide & 50.21 & 17.76 & 49.94 & 48.87 & 53.83 & 58.52 & 58.92 & \textbf{71.67} & \underline{ 69.56} & 58.37 \\
    Midjourney & 51.92 & 16.03 & 50.35 & 51.47 & 49.06 & \underline{ 62.26} & 53.67 & \textbf{69.56} & 50.40 & 57.87 \\
    SDv1.4 & 51.42 & 15.18 & 49.93 & 50.98 & 48.28 & 59.24 & 50.80 & \underline{ 74.48} & 51.17 & \textbf{81.39} \\
    SDv1.5 & 51.66 & 14.91 & 49.94 & 50.76 & 47.81 & 58.82 & 49.36 & \underline{ 74.39} & 51.04 & \textbf{81.01} \\
    VQDM & 50.24 & 17.01 & 49.96 & 49.48 & 51.86 & 57.96 & 52.56 & \underline{ 76.06} & \textbf{81.45} & 75.30 \\
    wukong & 50.31 & 16.66 & 49.97 & 49.92 & 51.15 & 56.65 & 52.36 & \underline{ 69.62} & 54.64 & \textbf{78.74} \\
    DALLE2 & 48.00 & 20.60 & 49.95 & 47.30 & 62.90 & 60.95 & 58.75 & \underline{ 65.75} & 51.40 & \textbf{73.40} \\
    SDXL & 61.25 & 24.88 & 60.30 & 56.69 & 59.03 & \underline{ 78.28} & 52.63 & \textbf{90.70} & 50.75 & 77.68 \\
    \hline
    Average & 58.99 & 35.19 & 50.61 & 60.09 & 63.28 & 61.89 & 55.89 & 63.94 & \underline{ 69.69} & \textbf{78.34} \\
    \bottomrule
    \end{tabular}%
    \caption{The detection accuracy (Downsampling) comparison between our approach and baselines.}
  \label{tab_map_downampling}%
\end{table*}%

\begin{table*}[]
  \centering
  
  \small
    \begin{tabular}{ccccccccccc}
    \toprule
        Generator & CNNSpot & FreDect & Fusing & GramNet & LNP   & LGrad & DIRE-G & DIRE-D & UnivFD & Ours \\
        \hline
    ProGAN & 99.27 & 81.83 & 50.69 & 98.94 & 94.12 & 97.21 & 77.93 & 50.18 & \underline{ 99.35} & \textbf{100.00} \\
    StyleGan & 90.80 & 72.00 & 47.53 & 90.25 & \underline{ 93.49} & 91.79 & 80.44 & 56.19 & 90.72 & \textbf{99.17} \\
    BigGAN & 61.36 & 57.93 & 51.18 & 61.15 & \underline{ 75.77} & 64.29 & 52.73 & 52.25 & \textbf{84.65} & 72.34 \\
    CycleGAN & 77.25 & 54.03 & 50.74 & 82.52 & 71.89 & 58.24 & 60.26 & 46.40 & \textbf{96.34} & \underline{ 92.49} \\
    StarGAN & 93.20 & 73.29 & 53.17 & 95.17 & \underline{ 99.50} & 97.29 & 69.22 & 47.37 & 95.72 & \textbf{100.00} \\
    GauGAN & \underline{ 85.10} & 54.90 & 49.60 & 82.65 & 52.67 & 57.91 & 45.73 & 40.67 & \textbf{97.51} & 67.57 \\
    Stylegan2 & 91.65 & 69.83 & 45.63 & 91.46 & \underline{ 97.11} & 94.44 & 77.45 & 70.22 & 87.12 & \textbf{99.17} \\
    whichfaceisreal & 84.73 & 49.15 & 45.04 & 86.02 & 79.90 & 61.07 & 51.34 & 51.33 & \underline{ 89.00} & \textbf{91.22} \\
    ADM & 61.52 & 32.85 & 33.52 & 57.21 & 76.01 & 59.18 & 60.45 & \textbf{99.62} & \underline{ 91.77} & 81.07 \\
    Glide & 60.63 & 33.03 & 40.11 & 55.83 & 58.44 & 65.90 & 71.15 & \textbf{99.53} & \underline{ 91.14} & 65.03 \\
    Midjourney & 54.47 & 31.72 & \underline{ 71.05} & 54.78 & 48.75 & 67.83 & 58.95 & \textbf{99.18} & 52.57 & 63.24 \\
    SDv1.4 & 55.50 & 31.56 & 39.14 & 53.80 & 47.75 & 62.72 & 51.93 & \textbf{99.71} & 62.25 & \underline{ 91.01} \\
    SDv1.5 & 55.97 & 31.51 & 39.22 & 53.68 & 47.63 & 63.31 & 50.42 & \textbf{99.71} & 62.56 & \underline{ 91.10} \\
    VQDM & 59.25 & 33.23 & 39.68 & 55.79 & 54.33 & 62.63 & 57.93 & \textbf{99.54} & \underline{ 95.71} & 84.03 \\
    wukong & 51.57 & 31.76 & 39.94 & 50.39 & 52.09 & 61.21 & 52.31 & \textbf{99.49} & 72.80 & \underline{ 87.03} \\
    DALLE2 & 55.21 & 34.84 & 45.38 & 50.55 & 72.05 & 67.48 & 69.84 & \textbf{99.58} & 72.99 & \underline{ 85.78} \\
    SDXL & 75.95 & 33.46 & 72.47 & 55.45 & 58.05 & 85.90 & 52.00 & \textbf{99.05} & 60.41 & \underline{ 85.01} \\
    \hline
    Average & 71.38 & 47.47 & 47.89 & 69.16 & 69.39 & 71.67 & 61.18 & 77.06 & \underline{ 82.51} & \textbf{85.60} \\
    \bottomrule
    \end{tabular}%
    \caption{The average precision (Downsampling) comparison between our approach and baselines.}
  \label{tab_map_downampling}%
\end{table*}%

\begin{table*}[]
  \centering
  
  \small
    \begin{tabular}{ccccccccccc}
    \toprule
        Generator & CNNSpot & FreDect & Fusing & GramNet & LNP   & LGrad & DIRE-G & DIRE-D & UnivFD & Ours \\
        \hline
    ProGAN & \textbf{99.95} & 85.02 & \underline{ 99.94} & 99.90 & 76.56 & 95.46 & 85.85 & 51.84 & 98.65 & 99.01 \\
    StyleGan & 83.32 & 80.35 & 83.15 & 84.84 & 68.36 & \underline{ 85.26} & 72.79 & 50.75 & 71.99 & \textbf{90.38} \\
    BigGAN & 68.03 & 71.90 & \underline{ 74.00} & 67.42 & 53.78 & 63.90 & 57.35 & 51.28 & \textbf{76.92} & 63.00 \\
    CycleGAN & 85.65 & 69.91 & 83.61 & \underline{ 86.41} & 52.65 & 53.48 & 65.44 & 50.34 & \textbf{94.66} & 75.47 \\
    StarGAN & 87.12 & 84.67 & \textbf{95.82} & \underline{ 92.95} & 63.16 & 92.25 & 80.55 & 41.52 & 89.62 & 78.71 \\
    GauGAN & 79.30 & 58.99 & \underline{ 81.08} & 72.98 & 49.23 & 61.09 & 62.72 & 39.23 & \textbf{97.46} & 60.65 \\
    Stylegan2 & 84.04 & 79.79 & \underline{ 88.43} & 87.49 & 76.75 & 86.09 & 63.20 & 51.04 & 62.11 & \textbf{91.99} \\
    whichfaceisreal & \textbf{82.70} & 49.60 & 69.70 & \underline{ 82.40} & 52.05 & 53.35 & 61.15 & 52.80 & 58.55 & 62.30 \\
    ADM & 59.30 & 65.81 & 55.83 & 57.88 & 70.30 & \underline{ 70.99} & 65.63 & \textbf{93.13} & 64.50 & 69.58 \\
    Glide & 55.28 & 67.34 & 53.27 & 54.45 & \underline{ 83.31} & 78.54 & 75.14 & \textbf{92.44} & 60.88 & 72.52 \\
    Midjourney & 51.37 & 51.60 & 51.30 & 50.68 & 68.52 & \underline{ 77.43} & 55.44 & \textbf{90.42} & 55.48 & 76.28 \\
    SDv1.4 & 51.23 & 49.10 & 48.70 & 52.87 & 64.31 & 62.93 & 47.04 & \textbf{92.87} & 54.56 & \underline{ 78.85} \\
    SDv1.5 & 51.46 & 48.89 & 49.07 & 52.94 & 64.17 & 63.32 & 47.21 & \textbf{92.87} & 54.61 & \underline{ 78.61} \\
    VQDM & 55.56 & 67.43 & 54.67 & 54.35 & 57.05 & 63.28 & 57.98 & \textbf{93.43} & \underline{ 76.47} & 70.53 \\
    wukong & 50.62 & 46.03 & 49.53 & 51.43 & 60.68 & 60.09 & 51.63 & \textbf{92.83} & 58.59 & \underline{ 74.23} \\
    DALLE2 & 49.30 & 75.55 & 51.35 & 49.10 & \underline{ 85.90} & 80.20 & 74.85 & \textbf{90.60} & 49.95 & 72.00 \\
    SDXL & 56.53 & 59.30 & 55.73 & 59.25 & 73.25 & \underline{ 79.53} & 54.20 & \textbf{92.08} & 50.78 & 76.50 \\
    \hline
    Average & 67.69 & 65.37 & 67.36 & 68.08 & 65.88 & \underline{ 72.19} & 63.42 & 71.73 & 69.16 & \textbf{75.92} \\
    \bottomrule
    \end{tabular}%
    \caption{The detection accuracy (Blur) comparison between our approach and baselines.}
  \label{tab_map_Blur}%
\end{table*}%

\begin{table*}[]
  \centering
  
  \small
    \begin{tabular}{ccccccccccc}
    \toprule
        Generator & CNNSpot & FreDect & Fusing & GramNet & LNP   & LGrad & DIRE-G & DIRE-D & UnivFD & Ours \\
        \hline
    ProGAN & \textbf{100.00} & 95.28 & \textbf{100.00} & \textbf{100.00} & 87.11 & 99.20 & 95.01 & 51.10 & 99.92 & \underline{ 99.98} \\
    StyleGan & \textbf{98.97} & 88.29 & 98.69 & \underline{ 98.70} & 79.52 & 96.50 & 84.15 & 52.58 & 92.84 & 97.59 \\
    BigGAN & 79.50 & 80.07 & \underline{ 83.19} & 79.82 & 54.31 & 69.38 & 59.17 & 51.23 & \textbf{91.98} & 64.41 \\
    CycleGAN & 91.80 & 76.50 & 90.70 & \underline{ 94.84} & 52.52 & 57.65 & 71.58 & 51.60 & \textbf{98.95} & 80.92 \\
    StarGAN & 97.92 & 93.58 & \underline{ 99.69} & 98.81 & 88.55 & \textbf{99.78} & 82.09 & 42.96 & 96.00 & 98.21 \\
    GauGAN & \underline{ 87.61} & 61.68 & 86.76 & 84.55 & 47.07 & 63.91 & 55.99 & 39.11 & \textbf{99.67} & 65.98 \\
    Stylegan2 & 99.00 & 88.60 & \textbf{99.60} & \underline{ 99.01} & 88.05 & 97.35 & 77.42 & 54.18 & 90.33 & 98.08 \\
    whichfaceisreal & 87.87 & 49.99 & \underline{ 90.48} & \textbf{91.14} & 49.75 & 57.24 & 61.85 & 58.79 & 68.65 & 75.42 \\
    ADM & 69.64 & 74.58 & 66.12 & 68.40 & 79.53 & \underline{ 80.35} & 68.50 & \textbf{99.66} & 80.74 & 78.88 \\
    Glide & 63.81 & 76.10 & 60.85 & 62.91 & \underline{ 92.53} & 86.81 & 80.90 & \textbf{98.98} & 77.25 & 83.25 \\
    Midjourney & 55.45 & 48.82 & 56.30 & 55.03 & 75.00 & 86.39 & 58.39 & \textbf{97.62} & 65.53 & \underline{ 88.03} \\
    SDv1.4 & 56.58 & 45.05 & 52.16 & 58.99 & 68.81 & 71.72 & 46.01 & \textbf{99.39} & 67.54 & \underline{ 90.80} \\
    SDv1.5 & 56.90 & 45.06 & 52.01 & 59.65 & 68.91 & 72.31 & 46.05 & \textbf{99.51} & 67.24 & \underline{ 90.65} \\
    VQDM & 64.46 & 76.80 & 63.89 & 62.06 & 62.44 & 70.94 & 59.59 & \textbf{99.87} & \underline{ 90.23} & 78.05 \\
    wukong & 54.09 & 38.48 & 51.90 & 55.64 & 63.56 & 67.87 & 52.51 & \textbf{99.47} & 74.59 & \underline{ 83.75} \\
    DALLE2 & 49.84 & 91.22 & 53.86 & 47.65 & \underline{ 95.03} & 88.19 & 81.81 & \textbf{99.60} & 54.28 & 90.81 \\
    SDXL & 77.49 & 65.50 & 67.66 & 56.84 & 75.09 & 86.39 & 54.02 & \textbf{99.25} & 59.07 & \underline{ 92.44} \\
    \hline
    Average & 75.94 & 70.33 & 74.93 & 74.94 & 72.22 & 79.53 & 66.77 & 76.17 & \underline{ 80.87} & \textbf{85.72} \\
    \bottomrule
    \end{tabular}%
    \caption{The average precision (Blur) comparison between our approach and baselines.}
  \label{tab_map_Blur}%
\end{table*}%

\end{document}


\maketitle

Our supplement consists of five parts, i.e., the parameters of high-pass filters, the structure of the classifier, the details experimental results of detection performance, complementary ablation studies, and AI-generated fake image visualization.

\begin{figure}[ht]
	\centering
  \includegraphics[width=3.2in,clip,trim=70 165 610 70]{Fig/fig 5}
	\caption{The specific kernel parameters of high-pass filters. }\label{fig_hpf}
\end{figure}

\section{High-pass Filters}
In our manuscript, we adopt 30 high-pass filters which are proposed in the image steganalysis domain. Various filters can project spatial images in different directions and extract diverse high frequency signals. Fig. \ref{fig_hpf} shows the specific parameters of high-pass filters. We show 7 kernel parameters for concision. The entire kernel parameters are as follows. Eight different variants of (a) by rotating (a) following eight directions \{ $\nearrow, \rightarrow, \searrow, \downarrow, \swarrow \leftarrow, \nwarrow, \uparrow $ \}. (b) is similar to (a). Four different variants of (c) by rotating (c) following four directions \{ $ \rightarrow,  \downarrow, \nearrow ,\searrow  $ \} (\{ $ \leftarrow ,\uparrow ,\swarrow ,\nwarrow  $ \} are the same as their reverse directions therefore we ignore them.). Four different variants of (d) by rotating (d) following four directions \{ $ \rightarrow,  \downarrow, \leftarrow, \uparrow $ \}. (e) is similar to (d). Therefore, we obtain 2*8+1*4+2*4+2=30 high-pass filters.

\section{Classifier Structure}

Since our proposed fingerprint exists across various AI-generated fake images, a simple cascaded CNN network can achieve satisfactory detection performance. The structure of our classifier is shown in Table \ref{tab_classifier}.

\begin{table}[htbp]
  \centering
    \begin{tabular}{cccc}
    \toprule
    Type  & Kernel num & With BN & Activation \\
    \hline
    Convo. & 32    & TRUE  & ReLU \\
    Convo. & 32    & TRUE  & ReLU \\
    Convo. & 32    & TRUE  & ReLU \\
    Convo. & 32    & TRUE  & ReLU \\
    Avg Pooling & None  & None  & None \\
    Convo. & 32    & TRUE  & ReLU \\
    Convo. & 32    & TRUE  & ReLU \\
    Avg Pooling & None  & None  & None \\
    Convo. & 32    & TRUE  & ReLU \\
    Convo. & 32    & TRUE  & ReLU \\
    Avg Pooling & None  & None  & None \\
    Convo. & 32    & TRUE  & ReLU \\
    Convo. & 32    & TRUE  & ReLU \\
    AdpAvgPool & None  & None  & None \\
    Flatten & None  & None  & None \\
    FC    & None  & FALSE & None \\
    \bottomrule
    \end{tabular}%
    \caption{The structure of the classifier.}
  \label{tab_classifier}%
\end{table}%

\section{Performance Evaluation}
In the manuscript, we adopt accuracy and average precision as the criterion for AI-generated fake image detection. Table \ref{tab_acc_nodistortion} -\ref{tab_map_Blur} shows the concrete experimental results. 

\section{Complementary Ablation Studies}

\begin{table}[htbp]
  \centering
  
    \begin{tabular}{ccccc}
    \toprule
          & \multirow{2}[0]{*}{w/o HPF} & \multicolumn{2}{c}{Patch Size} & \multirow{2}[0]{*}{Ours} \\
    Generator &       & 16    & 64    &  \\
    \hline
    ProGAN & 100.00  & 100.00  & 99.99  & 100.00  \\
    StyleGAN & 88.35  & 92.82  & 92.56  & 92.77  \\
    BigGAN & 79.35  & 84.78  & 94.25  & 95.80  \\
    CycleGAN & 69.57  & 54.16  & 68.02  & 70.17  \\
    StarGAN & 78.21  & 77.91  & 99.20  & 99.97  \\
    GauGAN & 64.26  & 50.71  & 68.50  & 71.58  \\
    Stylegan2 & 93.41  & 92.44  & 89.43  & 89.55  \\
    whichfaceisreal & 65.85  & 82.70  & 89.80  & 85.80  \\
    ADM   & 63.69  & 57.93  & 72.42  & 82.17  \\
    Glide & 71.03  & 84.99  & 79.77  & 83.79  \\
    Midjourney & 72.26  & 83.40  & 90.38  & 90.12  \\
    SDv1.4 & 87.35  & 92.78  & 95.71  & 95.38  \\
    SDv1.5 & 87.24  & 92.88  & 95.91  & 95.30  \\
    VQDM  & 69.71  & 75.76  & 89.83  & 88.91  \\
    Wukong & 78.82  & 86.90  & 89.88  & 91.07  \\
    DALLE2 & 80.85  & 96.35  & 93.80  & 96.60  \\
    Average & 78.12  & 81.66  & 88.09  & 89.31  \\
    \hline
    Degradation & \textcolor{gray}{-11.19}  & \textcolor{gray}{-7.65}  & \textcolor{gray}{-1.22}  \\
    \bottomrule
    \end{tabular}%
    \caption{The ablation studies of high-pass filters and the patch size.}
  \label{tab_ablation}%
\end{table}%

The size of rich/poor texture reconstructed images is based on the patch size and the number of patches. In the above experiments, we set the size of rich/poor texture region image as $256\times256$, and the size of each patch is $32\times 32$. In other words, each texture region image is made up of 64 patches. In the ablation experiments, we analyze the impact of patch size on the final performance. We set the size of the rich/poor patch as $16\times 16$ or $64\times 64$. The number of the reconstructed image is 256 or 16, respectively. The experimental results of the different patch sizes are shown in Table \ref{tab_ablation}. Our approach achieves the best performance with $32\times32$ patch size. The detection performance drops with a small patch size.

An alternative important component of fingerprint feature extraction is high-pass filters. Previous studies show that generated images exhibit more artifacts in the high frequency domain. Therefore, we adopt high-pass filters to magnify fake image artifacts and make the detector focus on the inter-pixel correlations. Based on Table \ref{tab_ablation} (w/o HPF), high-pass filters can improve our detection accuracy by 11\% for average cases.

\section{AI-generated fake image visualization}

\begin{figure*}[ht]
	\centering
  \includegraphics[width=6.8in,clip,trim=250 390 360 0]{Fig/fig 6}
	\caption{The visualization results of AI-generated fake images.}
\end{figure*}

In this supplement, we visualize various generative models including GAN-based models, Diffusion-based models, and their variants. Modern diffusion-based generator like Midjourney can create realistic-look synthetic images, which is hardly distinguished from real ones by humans.

\begin{table*}[]
  \centering
  
  \small
    \begin{tabular}{ccccccccccc}
    \toprule
    Generator & CNNSpot & FreDect & Fusing & GramNet & LNP   & LGrad & DIRE-G & DIRE-D & UnivFD & Ours \\
    \hline
    ProGAN & 100.00  & 99.36  & 100.00  & 99.99  & 99.67  & 99.83  & 95.19  & 52.75  & 99.81  & 100.00  \\
    StyleGAN & 90.17  & 78.02  & 85.20  & 87.05  & 91.75  & 91.08  & 83.03  & 51.31  & 84.93  & 92.77  \\
    BigGAN & 71.17  & 81.97  & 77.40  & 67.33  & 77.75  & 85.62  & 70.12  & 49.70  & 95.08  & 95.80  \\
    CycleGAN & 87.62  & 78.77  & 87.00  & 86.07  & 84.10  & 86.94  & 74.19  & 49.58  & 98.33  & 70.17  \\
    StarGAN & 94.60  & 94.62  & 97.00  & 95.05  & 99.92  & 99.27  & 95.47  & 46.72  & 95.75  & 99.97  \\
    GauGAN & 81.42  & 80.57  & 77.00  & 69.35  & 75.39  & 78.46  & 67.79  & 51.23  & 99.47  & 71.58  \\
    StyleGAN2 & 86.91  & 66.19  & 83.30  & 87.28  & 94.64  & 85.32  & 75.31  & 51.72  & 74.96  & 89.55  \\
    whichfaceisreal & 91.65  & 50.75  & 66.80  & 86.80  & 70.85  & 55.70  & 58.05  & 53.30  & 86.90  & 85.80  \\
    ADM   & 60.39  & 63.42  & 49.00  & 58.61  & 84.73  & 67.15  & 75.78  & 98.25  & 66.87  & 82.17  \\
    Glide & 58.07  & 54.13  & 57.20  & 54.50  & 80.52  & 66.11  & 71.75  & 92.42  & 62.46  & 83.79  \\
    Midjourney & 51.39  & 45.87  & 52.20  & 50.02  & 65.55  & 65.35  & 58.01  & 89.45  & 56.13  & 90.12  \\
    SDv1.4 & 50.57  & 38.79  & 51.00  & 51.70  & 85.55  & 63.02  & 49.74  & 91.24  & 63.66  & 95.38  \\
    SDv1.5 & 50.53  & 39.21  & 51.40  & 52.16  & 85.67  & 63.67  & 49.83  & 91.63  & 63.49  & 95.30  \\
    VQDM  & 56.46  & 77.80  & 55.10  & 52.86  & 74.46  & 72.99  & 53.68  & 91.90  & 85.31  & 88.91  \\
    Wukong & 51.03  & 40.30  & 51.70  & 50.76  & 82.06  & 59.55  & 54.46  & 90.90  & 70.93  & 91.07  \\
    DALLE2 & 50.45  & 34.70  & 52.80  & 49.25  & 88.75  & 65.45  & 66.48  & 92.45  & 50.75  & 96.60  \\
    Average & 70.78  & 64.03  & 68.38  & 68.67  & 83.84  & 75.34  & 68.68  & 71.53  & 78.43  & 89.31  \\
    \bottomrule
    \end{tabular}%
    \caption{The detection accuracy (no distortion) comparison between our approach and baselines.}
  \label{tab_acc_nodistortion}%
\end{table*}%

\begin{table*}[]
  \centering
  
  \small
    \begin{tabular}{ccccccccccc}
    \toprule
    Generator & CNNSpot & FreDect & Fusing & GramNet & LNP   & LGrad & DIRE-G & DIRE-D & UnivFD & Ours \\
    \hline
    ProGAN & 100.00  & 99.99  & 100.00  & 100.00  & 99.99  & 100.00  & 99.08  & 58.79  & 100.00  & 100.00  \\
    StyleGAN & 99.83  & 88.98  & 99.50  & 99.23  & 98.60  & 98.31  & 91.74  & 56.68  & 97.56  & 98.96  \\
    BigGAN & 85.99  & 93.62  & 90.70  & 81.79  & 84.32  & 92.93  & 75.25  & 46.91  & 99.27  & 99.42  \\
    CycleGAN & 94.94  & 84.78  & 95.50  & 95.33  & 92.83  & 95.01  & 80.56  & 50.03  & 99.80  & 85.26  \\
    StarGAN & 99.04  & 99.49  & 99.80  & 99.23  & 100.00  & 100.00  & 99.34  & 40.64  & 99.37  & 100.00  \\
    GauGAN & 90.82  & 82.84  & 88.30  & 84.99  & 78.85  & 95.43  & 72.15  & 47.34  & 99.98  & 81.33  \\
    StyleGAN2 & 99.48  & 82.54  & 99.60  & 99.11  & 99.59  & 97.89  & 88.29  & 58.03  & 97.90  & 97.74  \\
    whichfaceisreal & 99.85  & 55.85  & 93.30  & 95.21  & 91.45  & 57.99  & 60.13  & 59.02  & 96.73  & 95.26  \\
    ADM   & 75.67  & 61.77  & 94.10  & 73.11  & 94.20  & 72.95  & 85.84  & 99.79  & 86.81  & 93.40  \\
    Glide & 72.28  & 52.92  & 77.50  & 66.76  & 88.86  & 80.42  & 78.35  & 99.54  & 83.81  & 94.04  \\
    Midjourney & 66.24  & 46.09  & 70.00  & 56.82  & 76.86  & 71.86  & 61.86  & 97.32  & 74.00  & 96.48  \\
    SDv1.4 & 61.20  & 37.83  & 65.40  & 59.83  & 94.31  & 62.37  & 49.87  & 98.61  & 86.14  & 99.06  \\
    SDv1.5 & 61.56  & 37.76  & 65.70  & 60.37  & 93.92  & 62.85  & 49.52  & 98.83  & 85.84  & 99.06  \\
    VQDM  & 68.83  & 85.10  & 75.60  & 61.13  & 87.35  & 77.47  & 54.57  & 98.98  & 96.53  & 96.26  \\
    Wukong & 57.34  & 39.58  & 64.60  & 55.62  & 92.38  & 62.48  & 55.38  & 98.37  & 91.07  & 97.54  \\
    DALLE2 & 53.51  & 38.20  & 68.12  & 49.82  & 96.14  & 82.55  & 74.48  & 99.71  & 63.04  & 99.56  \\
    Average & 80.41  & 67.96  & 84.23  & 77.40  & 91.85  & 81.91  & 73.53  & 75.54  & 91.12  & 95.72  \\
    \bottomrule
    \end{tabular}%
    \caption{The average precision (no distortion) comparison between our approach and baselines.}
  \label{tab_map_nodistortion}%
\end{table*}%

\begin{table*}[]
  \centering
  
  \small
    \begin{tabular}{ccccccccccc}
    \toprule
        Generator & CNNSpot & FreDect & Fusing & GramNet & LNP   & LGrad & DIRE-G & DIRE-D & UnivFD & Ours \\
        \hline
    ProGAN & 99.96  & 84.40  & 99.64  & 99.94  & 71.16  & 55.70  & 98.92  & 51.55  & 99.34  & 97.84  \\
    StyleGAN & 75.00  & 72.30  & 70.45  & 77.74  & 56.02  & 55.90  & 75.93  & 50.89  & 79.74  & 82.49  \\
    BigGAN & 62.05  & 62.90  & 63.50  & 62.28  & 51.20  & 51.35  & 67.85  & 51.08  & 88.22  & 65.25  \\
    CycleGAN & 83.16  & 71.35  & 80.96  & 86.98  & 57.27  & 57.15  & 74.68  & 50.53  & 96.14  & 71.99  \\
    StarGAN & 79.26  & 83.79  & 92.02  & 88.72  & 50.75  & 51.68  & 83.08  & 41.55  & 95.32  & 60.21  \\
    GauGAN & 69.89  & 64.99  & 66.91  & 65.69  & 50.06  & 49.70  & 77.84  & 39.31  & 98.61  & 73.71  \\
    StyleGAN2 & 71.29  & 73.34  & 60.19  & 78.87  & 58.81  & 55.99  & 68.99  & 51.36  & 69.41  & 82.71  \\
    whichfaceisreal & 79.10  & 50.25  & 59.90  & 86.40  & 50.12  & 52.90  & 59.05  & 54.35  & 70.20  & 79.40  \\
    ADM   & 51.28  & 67.72  & 50.01  & 51.62  & 51.28  & 46.50  & 62.49  & 91.92  & 64.41  & 62.64  \\
    Glide & 51.92  & 66.13  & 51.99  & 51.59  & 50.97  & 45.09  & 73.11  & 91.38  & 64.14  & 68.01  \\
    Midjourney & 50.90  & 56.61  & 50.85  & 49.85  & 51.60  & 62.32  & 58.90  & 90.78  & 55.43  & 57.87  \\
    SDv1.4 & 49.82  & 52.95  & 50.16  & 49.44  & 52.66  & 53.87  & 48.40  & 92.53  & 56.48  & 75.00  \\
    SDv1.5 & 49.90  & 52.71  & 50.13  & 49.56  & 52.31  & 53.62  & 48.48  & 92.56  & 56.19  & 74.87  \\
    VQDM  & 51.24  & 76.51  & 51.61  & 50.98  & 50.98  & 44.06  & 54.82  & 92.47  & 78.89  & 64.94  \\
    Wukong & 50.01  & 52.60  & 50.50  & 49.62  & 51.55  & 52.49  & 50.88  & 92.14  & 62.75  & 67.91  \\
    DALLE2 & 49.70  & 82.70  & 50.00  & 48.20  & 50.50  & 36.40  & 60.40  & 89.89  & 50.35  & 70.35  \\
    Average & 64.03  & 66.95  & 62.43  & 65.47  & 53.58  & 51.55  & 66.49  & 70.27  & 74.10  & 72.48  \\
    \bottomrule
    \end{tabular}%
    \caption{The detection accuracy (JPG compression) comparison between our approach and baselines.}
  \label{tab_acc_jpg}%
\end{table*}%

\begin{table*}[]
  \centering
  
  \small
    \begin{tabular}{ccccccccccc}
    \toprule
        Generator & CNNSpot & FreDect & Fusing & GramNet & LNP   & LGrad & DIRE-G & DIRE-D & UnivFD & Ours \\
        \hline
        ProGAN & 100.00  & 95.45  & 100.00  & 100.00  & 92.64  & 68.65  & 99.94  & 50.95  & 99.98  & 99.73 \\
    StyleGAN & 98.60  & 79.65  & 98.62  & 98.24  & 75.36  & 75.48  & 94.12  & 53.59  & 94.90  & 91.12 \\
    BigGAN & 88.36  & 69.81  & 91.39  & 83.14  & 65.83  & 54.21  & 78.07  & 50.21  & 97.65  & 69.41 \\
    CycleGAN & 96.08  & 82.42  & 97.13  & 96.32  & 82.86  & 70.26  & 84.29  & 52.13  & 99.47  & 83.3 \\
    StarGAN & 93.63  & 94.37  & 99.29  & 97.16  & 68.50  & 61.44  & 92.16  & 39.44  & 99.07  & 70.29 \\
    GauGAN & 95.53  & 68.76  & 96.69  & 89.54  & 57.67  & 51.42  & 76.22  & 37.73  & 99.91  & 82.41 \\
    StyleGAN2 & 98.08  & 82.48  & 98.26  & 98.42  & 85.29  & 77.57  & 92.02  & 54.50  & 94.65  & 91.03 \\
    whichfaceisreal & 94.34  & 54.14  & 95.27  & 95.19  & 69.25  & 55.02  & 61.86  & 60.36  & 88.13  & 89.49 \\
    ADM   & 64.18  & 87.14  & 88.56  & 60.64  & 60.73  & 44.73  & 71.20  & 98.99  & 84.00  & 75.07 \\
    Glide & 69.60  & 91.06  & 80.01  & 62.67  & 60.94  & 42.16  & 83.60  & 98.44  & 85.18  & 81.04 \\
    Midjourney & 61.95  & 72.61  & 75.21  & 54.79  & 65.18  & 68.04  & 65.10  & 97.96  & 71.41  & 66.42 \\
    SDv1.4 & 55.52  & 66.99  & 64.10  & 52.56  & 68.26  & 56.03  & 45.14  & 99.15  & 75.23  & 86.47 \\
    SDv1.5 & 56.47  & 66.15  & 64.28  & 53.32  & 67.23  & 56.30  & 45.22  & 99.19  & 74.82  & 86.59 \\
    VQDM  & 64.82  & 91.26  & 76.46  & 56.98  & 57.90  & 40.91  & 58.94  & 99.39  & 93.58  & 78.54 \\
    Wukong & 54.78  & 65.54  & 63.78  & 51.40  & 63.69  & 54.32  & 50.50  & 99.23  & 83.69  & 80.26 \\
    DALLE2 & 47.03  & 98.58  & 54.97  & 43.52  & 49.61  & 36.37  & 66.37  & 99.51  & 58.58  & 95.59 \\
    Average & 77.43  & 79.15  & 84.00  & 74.62  & 68.18  & 57.06  & 72.80  & 74.42  & 87.52  & 83.10  \\
    \bottomrule
    \end{tabular}%
    \caption{The average precision (JPG compression) comparison between our approach and baselines.}
  \label{tab_map_jpg}%
\end{table*}%

\begin{table*}[]
  \centering
  
  \small
    \begin{tabular}{ccccccccccc}
    \toprule
        Generator & CNNSpot & FreDect & Fusing & GramNet & LNP   & LGrad & DIRE-G & DIRE-D & UnivFD & Ours \\
        \hline
    ProGAN & 88.00  & 60.85  & 50.00  & 88.30  & 71.50  & 81.46  & 68.01  & 49.74  & 95.83  & 99.92  \\
    StyleGAN & 64.46  & 61.81  & 50.00  & 67.69  & 69.19  & 71.32  & 66.52  & 51.74  & 72.63  & 90.37  \\
    BigGAN & 52.02  & 52.70  & 50.00  & 52.20  & 61.60  & 58.23  & 52.08  & 50.68  & 73.00  & 72.35  \\
    CycleGAN & 60.83  & 48.75  & 50.00  & 67.41  & 67.71  & 52.42  & 57.53  & 46.21  & 89.21  & 83.76  \\
    StarGAN & 64.58  & 51.68  & 50.00  & 77.89  & 56.08  & 58.38  & 59.80  & 58.39  & 88.02  & 99.90  \\
    GauGAN & 65.61  & 50.55  & 50.00  & 63.97  & 49.73  & 55.04  & 44.82  & 52.61  & 91.23  & 62.07  \\
    StyleGAN2 & 65.69  & 63.50  & 50.00  & 69.40  & 74.46  & 69.93  & 66.79  & 60.05  & 68.66  & 89.00  \\
    whichfaceisreal & 76.50  & 49.70  & 50.00  & 79.80  & 55.05  & 56.70  & 50.85  & 50.00  & 73.75  & 79.55  \\
    ADM   & 50.09  & 15.68  & 49.98  & 49.38  & 53.97  & 55.89  & 54.63  & 75.27  & 71.94  & 71.12  \\
    Glide & 50.21  & 17.76  & 49.94  & 48.87  & 48.52  & 58.52  & 58.92  & 71.67  & 69.56  & 58.37  \\
    Midjourney & 51.92  & 16.03  & 50.35  & 51.47  & 54.00  & 62.26  & 53.67  & 69.56  & 50.40  & 57.87  \\
    SDv1.4 & 51.42  & 15.18  & 49.93  & 50.98  & 55.25  & 59.24  & 50.80  & 74.48  & 51.17  & 81.39  \\
    SDv1.5 & 51.66  & 14.91  & 49.94  & 50.76  & 55.15  & 58.82  & 49.36  & 74.39  & 51.04  & 81.01  \\
    VQDM  & 50.24  & 17.01  & 49.96  & 49.48  & 46.14  & 57.96  & 52.56  & 76.06  & 81.45  & 75.30  \\
    Wukong & 50.31  & 16.66  & 49.97  & 49.92  & 57.58  & 56.65  & 52.36  & 69.62  & 54.64  & 78.74  \\
    DALLE2 & 48.00  & 20.60  & 49.95  & 47.30  & 45.30  & 60.95  & 58.75  & 65.75  & 51.40  & 73.40  \\
    Average & 58.85  & 35.84  & 50.00  & 60.30  & 57.58  & 60.86  & 56.09  & 62.26  & 70.87  & 78.36  \\
    \bottomrule
    \end{tabular}%
    \caption{The detection accuracy (Downsampling) comparison between our approach and baselines.}
  \label{tab_map_downampling}%
\end{table*}%

\begin{table*}[]
  \centering
  
  \small
    \begin{tabular}{ccccccccccc}
    \toprule
        Generator & CNNSpot & FreDect & Fusing & GramNet & LNP   & LGrad & DIRE-G & DIRE-D & UnivFD & Ours \\
        \hline
    ProGAN & 99.27  & 81.83  & 50.69  & 98.94  & 92.36  & 97.21  & 77.93  & 50.18  & 99.35  & 100.00  \\
    StyleGAN & 90.80  & 72.00  & 47.53  & 90.25  & 93.60  & 91.79  & 80.44  & 56.19  & 90.72  & 99.17  \\
    BigGAN & 61.36  & 57.93  & 51.18  & 61.15  & 73.61  & 64.29  & 52.73  & 52.25  & 84.65  & 72.34  \\
    CycleGAN & 77.25  & 54.03  & 50.74  & 82.52  & 78.42  & 58.24  & 60.26  & 46.40  & 96.34  & 92.49  \\
    StarGAN & 93.20  & 73.29  & 53.17  & 95.17  & 99.30  & 97.29  & 69.22  & 47.37  & 95.72  & 100.00  \\
    GauGAN & 85.10  & 54.90  & 49.60  & 82.65  & 50.30  & 57.91  & 45.73  & 40.67  & 97.51  & 67.57  \\
    StyleGAN2 & 91.65  & 69.83  & 45.63  & 91.46  & 96.17  & 94.44  & 77.45  & 70.22  & 87.12  & 99.17  \\
    whichfaceisreal & 84.73  & 49.15  & 45.04  & 86.02  & 68.15  & 61.07  & 51.34  & 51.33  & 89.00  & 91.22  \\
    ADM   & 61.52  & 32.85  & 33.52  & 57.21  & 59.82  & 59.18  & 60.45  & 99.62  & 91.77  & 81.07  \\
    Glide & 60.63  & 33.03  & 40.11  & 55.83  & 47.40  & 65.90  & 71.15  & 99.53  & 91.14  & 65.03  \\
    Midjourney & 54.47  & 31.72  & 71.05  & 54.78  & 55.27  & 67.83  & 58.95  & 99.18  & 52.57  & 63.24  \\
    SDv1.4 & 55.50  & 31.56  & 39.14  & 53.80  & 58.72  & 62.72  & 51.93  & 99.71  & 62.25  & 91.01  \\
    SDv1.5 & 55.97  & 31.51  & 39.22  & 53.68  & 58.79  & 63.31  & 50.42  & 99.71  & 62.56  & 91.10  \\
    VQDM  & 59.25  & 33.23  & 39.68  & 55.79  & 47.85  & 62.63  & 57.93  & 99.54  & 95.71  & 84.03  \\
    Wukong & 51.57  & 31.76  & 39.94  & 50.39  & 61.90  & 61.21  & 52.31  & 99.49  & 72.80  & 87.03  \\
    DALLE2 & 55.21  & 34.84  & 45.38  & 50.55  & 47.53  & 67.48  & 69.84  & 99.58  & 72.99  & 85.78  \\
    Average & 71.09  & 48.34  & 46.35  & 70.01  & 68.08  & 70.78  & 61.75  & 75.69  & 83.89  & 85.55  \\
    \bottomrule
    \end{tabular}%
    \caption{The average precision (Downsampling) comparison between our approach and baselines.}
  \label{tab_map_downampling}%
\end{table*}%

\begin{table*}[]
  \centering
  
  \small
    \begin{tabular}{ccccccccccc}
    \toprule
        Generator & CNNSpot & FreDect & Fusing & GramNet & LNP   & LGrad & DIRE-G & DIRE-D & UnivFD & Ours \\
        \hline
    ProGAN & \textbf{99.95}  & 85.02  & 99.94  & 99.90  & 84.67  & 95.46  & 85.85  & 51.84  & 98.65  & 99.01  \\
    StyleGAN & 83.32  & 80.35  & 83.15  & 84.84  & 76.85  & 85.26  & 72.79  & 50.75  & 71.99  & \textbf{90.38}  \\
    BigGAN & 68.03  & 71.90  & 74.00  & 67.42  & 57.30  & 63.90  & 57.35  & 51.28  & \textbf{76.92}  & 63.00  \\
    CycleGAN & 85.65  & 69.91  & 83.61  & 86.41  & 54.39  & 53.48  & 65.44  & 50.34  & \textbf{94.66}  & 75.47  \\
    StarGAN & 87.12  & 84.67  & \textbf{95.82}  & 92.95  & 78.06  & 92.25  & 80.55  & 41.52  & 89.62  & 78.71  \\
    GauGAN & 79.30  & 58.99  & 81.08  & 72.98  & 52.00  & 61.09  & 62.72  & 39.23  & \textbf{97.46}  & 60.65  \\
    StyleGAN2 & 84.04  & 79.79  & 88.43  & 87.49  & 86.67  & 86.09  & 63.20  & 51.04  & 62.11  & \textbf{91.99}  \\
    whichfaceisreal & \textbf{82.70}  & 49.60  & 69.70  & 82.40  & 47.75  & 53.35  & 61.15  & 52.80  & 58.55  & 62.30  \\
    ADM   & 59.30  & 65.81  & 55.83  & 57.88  & \textbf{77.05}  & 70.99  & 65.63  & 93.13  & 64.50  & 69.58  \\
    Glide & 55.28  & 67.34  & 53.27  & 54.45  & 82.86  & 78.54  & 75.14  & \textbf{92.44}  & 60.88  & 72.52  \\
    Midjourney & 51.37  & 51.60  & 51.30  & 50.68  & 54.21  & 77.43  & 55.44  & \textbf{90.42}  & 55.48  & 76.28  \\
    SDv1.4 & 51.23  & 49.10  & 48.70  & 52.87  & 63.47  & 62.93  & 47.04  & \textbf{92.87}  & 54.56  & 78.85  \\
    SDv1.5 & 51.46  & 48.89  & 49.07  & 52.94  & 64.03  & 63.32  & 47.21  & \textbf{92.87}  & 54.61  & 78.61  \\
    VQDM  & 55.56  & 67.43  & 54.67  & 54.35  & 64.82  & 63.28  & 57.98  & \textbf{93.43}  & 76.47  & 70.53  \\
    Wukong & 50.62  & 46.03  & 49.53  & 51.43  & 62.64  & 60.09  & 51.63  & \textbf{92.83}  & 58.59  & 74.23  \\
    DALLE2 & 49.30  & 75.55  & 51.35  & 49.10  & 79.75  & 80.20  & 74.85  & \textbf{90.60}  & 49.95  & 72.00  \\
    Average & 68.39  & 65.75  & 68.09  & 68.63  & 67.91  & 71.73  & 64.00  & 70.46  & 70.31  & \textbf{75.99}  \\
    \bottomrule
    \end{tabular}%
    \caption{The detection accuracy (Blur) comparison between our approach and baselines.}
  \label{tab_map_Blur}%
\end{table*}%

\begin{table*}[]
  \centering
  
  \small
    \begin{tabular}{ccccccccccc}
    \toprule
        Generator & CNNSpot & FreDect & Fusing & GramNet & LNP   & LGrad & DIRE-G & DIRE-D & UnivFD & Ours \\
        \hline
    ProGAN & \textbf{100.00}  & 95.28  & \textbf{100.00}  & \textbf{100.00}  & 93.18  & 99.20  & 95.01  & 51.10  & 99.92  & 99.98  \\
    StyleGAN & \textbf{98.97}  & 88.29  & 98.69  & 98.70  & 88.32  & 96.50  & 84.15  & 52.58  & 92.84  & 97.59  \\
    BigGAN & 79.50  & 80.07  & 83.19  & 79.82  & 56.50  & 69.38  & 59.17  & 51.23  & \textbf{91.98}  & 64.41  \\
    CycleGAN & 91.80  & 76.50  & 90.70  & 94.84  & 59.92  & 57.65  & 71.58  & 51.60  & \textbf{98.95}  & 80.92  \\
    StarGAN & 97.92  & 93.58  & 99.69  & 98.81  & 96.68  & \textbf{99.78}  & 82.09  & 42.96  & 96.00  & 98.21  \\
    GauGAN & 87.61  & 61.68  & 86.76  & 84.55  & 49.20  & 63.91  & 55.99  & 39.11  & \textbf{99.67}  & 65.98  \\
    StyleGAN2 & 99.00  & 88.60  & \textbf{99.60}  & 99.01  & 95.04  & 97.35  & 77.42  & 54.18  & 90.33  & 98.08  \\
    whichfaceisreal & 87.87  & 49.99  & 90.48  & 91.14  & 43.83  & 57.24  & 61.85  & 58.79  & 68.65  & 75.42  \\
    ADM   & 69.64  & 74.58  & 66.12  & 68.40  & 86.83  & 80.35  & 68.50  & 99.66  & 80.74  & 78.88  \\
    Glide & 63.81  & 76.10  & 60.85  & 62.91  & 91.03  & 86.81  & 80.90  & 98.98  & 77.25  & 83.25  \\
    Midjourney & 55.45  & 48.82  & 56.30  & 55.03  & 57.70  & 86.39  & 58.39  & 97.62  & 65.53  & 88.03  \\
    SDv1.4 & 56.58  & 45.05  & 52.16  & 58.99  & 71.34  & 71.72  & 46.01  & 99.39  & 67.54  & 90.80  \\
    SDv1.5 & 56.90  & 45.06  & 52.01  & 59.65  & 72.51  & 72.31  & 46.05  & 99.51  & 67.24  & 90.65  \\
    VQDM  & 64.46  & 76.80  & 63.89  & 62.06  & 70.96  & 70.94  & 59.59  & 99.87  & 90.23  & 78.05  \\
    Wukong & 54.09  & 38.48  & 51.90  & 55.64  & 71.29  & 67.87  & 52.51  & 99.47  & 74.59  & 83.75  \\
    DALLE2 & 49.84  & 91.22  & 53.86  & 47.65  & 93.06  & 88.19  & 81.81  & 99.60  & 54.28  & 90.81  \\
    Average & 75.84  & 70.63  & 75.39  & 76.08  & 74.84  & 79.10  & 67.56  & 74.73  & 82.23  & 85.40  \\
    \bottomrule
    \end{tabular}%
    \caption{The average precision (Blur) comparison between our approach and baselines.}
  \label{tab_map_Blur}%
\end{table*}%

